\documentclass[10pt,twocolumn,letterpaper]{article}

\newif\ifarxiv
\arxivtrue % arxiv version
\ifarxiv
  \usepackage[pagenumbers]{cvpr}
  \newcommand{\ARXIVversion}[2]{#1}
\else
  \usepackage{cvpr}
  \errmessage{Update ARXIV paper number in the bib!}
  \newcommand{\ARXIVversion}[2]{#2}
\fi
\usepackage{graphicx}
\usepackage{amsmath}
\usepackage{amssymb}
\usepackage{booktabs}
\usepackage{stmaryrd}
\usepackage{soul}

\usepackage[pagebackref,breaklinks,colorlinks]{hyperref}

\usepackage{multirow}
\usepackage{multicol}
\usepackage{enumitem}
\usepackage{color}
\usepackage{xcolor}
\usepackage{flushend}

\DeclareMathOperator*{\argmax}{arg\,max}
\DeclareMathOperator*{\argmin}{arg\,min}

\newcommand{\datasetname}[1]{\textbf{ChangeIt}}
\newcommand{\paragraphcustom}[1]{\vspace{2pt}\noindent\textbf{#1}}
\newcommand{\hideme}[1]{}

\usepackage[capitalize]{cleveref}
\crefname{section}{Sec.}{Secs.}
\Crefname{section}{Section}{Sections}
\Crefname{table}{Table}{Tables}
\crefname{table}{Tab.}{Tabs.}

\def\cvprPaperID{3656} % *** Enter the CVPR Paper ID here
\def\confName{CVPR}
\def\confYear{2022}

\begin{document}


\title{Look for the Change: Learning Object States and \\ State-Modifying Actions from Untrimmed Web Videos}

\author{
	Tom\'{a}\v{s} Sou\v{c}ek\textsuperscript{1}%\footnotemark[1]
	\quad\quad\quad
	Jean-Baptiste Alayrac\textsuperscript{2}
	\quad\quad\quad
	Antoine Miech\textsuperscript{2}
	\\
	%	\quad\quad
	Ivan Laptev\textsuperscript{3}
	\quad\quad\quad%\quad\quad\quad\quad
	Josef Sivic\textsuperscript{1}
	\\
	\small{$^1$CIIRC CTU \quad $^2$DeepMind \quad $^3$ENS/Inria}
	\\
	\small{\texttt{tomas.soucek@cvut.cz}}
	\\
	\small{\url{https://data.ciirc.cvut.cz/public/projects/2022LookForTheChange/}}
}

% \and
\maketitle

%%%%%%%%% ABSTRACT
\begin{abstract}
Human actions often induce changes of object states such as ``cutting an apple”, ``cleaning shoes” or ``pouring coffee”. 
In this paper, we seek to temporally localize object states (e.g. ``empty” and ``full” cup) together with the corresponding state-modifying actions (``pouring coffee”) in long uncurated videos with minimal supervision.  The contributions of this work are threefold. 
First, we develop a self-supervised model for jointly learning state-modifying actions together with the corresponding object states from an uncurated set of videos from the Internet. The model is self-supervised by the causal ordering signal, i.e. initial object state $\rightarrow$ manipulating action $\rightarrow$ end state. 
Second,  to cope with noisy uncurated training data, our model incorporates a noise adaptive weighting module supervised by a small number of annotated still images, that allows to efficiently filter out irrelevant videos during training. 
Third, we collect a new dataset with more than 2600 hours of video and 34 thousand changes of object states, and manually annotate a part of this data to validate our approach.
Our results demonstrate substantial improvements over prior work in both action and object state-recognition in video. 
\end{abstract}

%%%%%%%%% BODY TEXT
%%%%%%%%%%%%%%%%%%%%

\footnotetext[1]{Czech Institute of Informatics, Robotics and Cybernetics at the Czech Technical University in Prague.}
\footnotetext[3]{D\'{e}partement d'informatique de l'ENS, \'{E}cole normale sup\'{e}rieure, CNRS, PSL Research University, 75005 Paris, France.}

\section{Introduction}\label{sec:intro}

\begin{figure}[t]
    \centering
    {\footnotesize
    \includegraphics[width=0.99\linewidth]{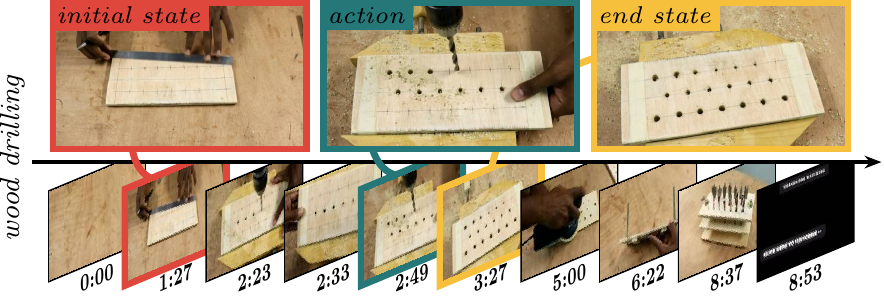}
    \vspace{5pt}\\
    \includegraphics[width=0.99\linewidth]{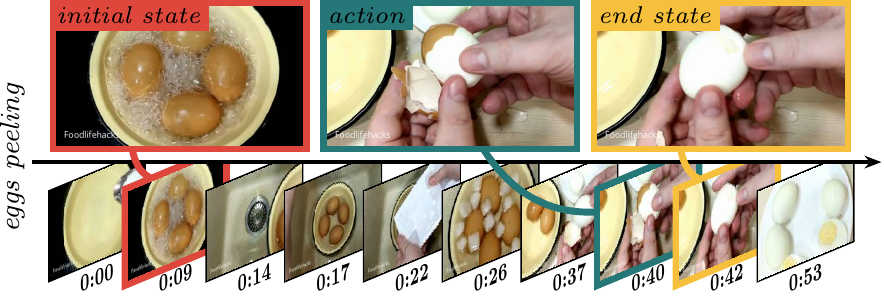}
    \vspace{-4pt}
    \caption{
    {\bf Examples of object states and state-modifying actions} learnt by our model from a dataset of long uncurated web videos. %lasting up to 15 minutes. 
    In each example the top row shows:  the  initial state in the video (left), the state-modifying action (middle), and the end-state (right). The bottom row shows video frames sampled from the entire video with their corresponding timestamps. It illustrates the difficulty of finding the correct temporal localization of the object states and the actions in the entire video.   
    }

    \label{fig:teaser}
    \vspace*{-0.3cm}
    }
\end{figure}

Human actions often induce changes of the state of an object, as illustrated in Figure~\ref{fig:teaser}. 
Examples include ``cutting an apple”, ``cleaning shoes”, ``tying a tie” or ``filling-up a cup with coffee”. 
People can easily recognize such actions and the resulting changes of object states~\cite{Brady06}, for example, when watching instructional videos. 
Furthermore, people can  reproduce the actions in their environment, \eg when following a recipe from a cooking video. 
However, artificial system with similar cognitive abilities is yet to be developed. 
Existing methods for recognizing object states and state-modifying actions address small-scale setups (5 objects and short manually curated videos)~\cite{Alayrac16ObjectStates} or controlled environments~\cite{dima2014youdo}. 
At the same time, progress on automatic understanding of causal relations between actions and object states in the wild would be a major step in embodied video understanding and robotics. 
However, the task is challenging given the large amount and variability of existing object-action pairs as well as the difficulty of manually collecting and annotating video data for it. 

In this paper, we investigate whether the learning of object states and corresponding state-modifying actions can be scaled-up to noisy uncurated videos from the web while using only minimal supervision. 
The contribution of this work is threefold as we outline below.

First, we develop a self-supervised model for jointly learning state-modifying actions and object states from an uncurated set of videos obtained from a video search engine. 
We explore the causal ordering in the video as a free supervisory signal and use it to discover the changing states of objects and state-modifying actions. We define it by the sequence of initial object state $\rightarrow$ manipulating action $\rightarrow$ end state, as illustrated in Figure~\ref{fig:teaser}.
While the prior work on this problem~\cite{Alayrac16ObjectStates} was limited to closed-form linear classifiers, our model is amenable to large-scale learning using stochastic gradient descent and supports non-linear multi-layer models.

Second, to cope with noisy uncurated data that may include a large proportion of irrelevant videos (\eg videos of Apple laptops when learning “cutting an apple”), our model incorporates a noise adaptive weighting module that allows to filter out irrelevant videos. This noise adaptive weighting module is supervised by a small number of still images depicting the two states of the object, which are easy to collect using currently available image search engines. This attention mechanism allows us to scale our method to noisy uncurated data, as we show by experimental results. 

Third, we collect a new ``\datasetname{}" dataset with more than 2600 hours of video and 34 thousand changes of object states. We manually annotate a portion of this data for evaluation. 
To validate our approach, we show results on this new uncurated dataset as well as on the existing smaller curated video dataset from~\cite{Alayrac16ObjectStates}. We ablate key components of our method and demonstrate substantial improvements over prior work both in action and object state localization. The dataset, the code, and a trained model are publicly available.%\footnote{\url{https://data.ciirc.cvut.cz/public/projects/2022LookForTheChange/}}

\begin{figure*}
  \centering
  \includegraphics[width=1.0\linewidth]{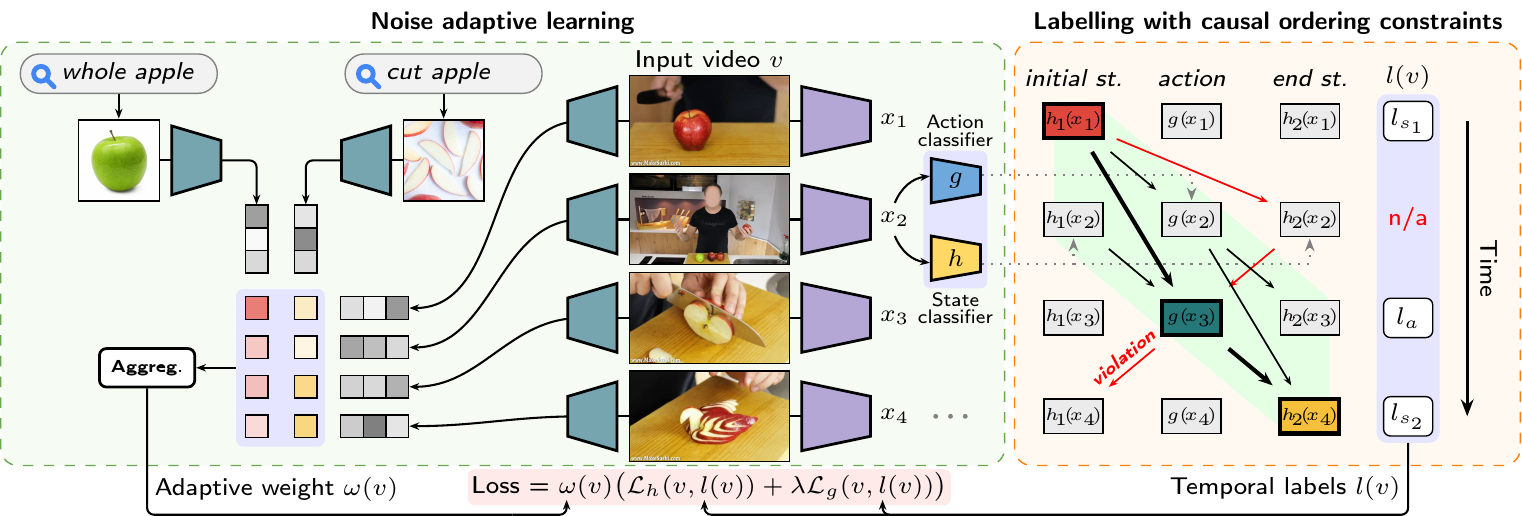}
  \vspace*{-5mm}
  \caption{{\bf Model overview.} Given a set of input {\em noisy untrimmed videos} from the web depicting a state-changing action (here \textit{cutting apple}) 
  our approach learns action classifier $g$ and object state classifier $h$ that output temporal labels $l$ of the input videos with temporal locations of {\em initial object state} $\rightarrow$ {\em manipulating action} $\rightarrow$ {\em end object state} that satisfy the {\em causal ordering constraint}. This is achieved by minimizing a new noise adaptive learning objective that downweights irrelevant videos with adaptive weight $\omega$ measuring similarity to a small number of exemplar images.
  The learning proceeds by iteratively \textbf{(i)} learning action and state classifiers, $g$ and $h$, given the current labels $l$ of the input videos and \textbf{(ii)} finding the labels~$l$ of the videos that respects the causal ordering constraints.
  }
  \vspace*{-3mm}
  \label{fig:overview}
\end{figure*}

%%%%%%%%%%%%%%%%%%%%%%

\section{Related work}\label{sec:relwork}

\noindent\textbf{Video and Language.}
A large body of work in automatic video understanding studies the use of natural language or speech data to train models for action and object state recognition.
Prior work~\cite{Alayrac2020MMVN,clip,chowdhury2018webly,dong19dual,gong14multi,gong14improving,miech19howto100m,miech2019end2end,pan16jointly,xu2015jointly,
wang2018learning,wray2019fine} leveraged image and video description datasets~\cite{lin14coco, miech19howto100m, plummer2015flickr30k, rohrbach17movie, xu16msrvtt, youcook2} to learn a joint vision-language embedding space, where visual and textual data are semantically aligned.
In particular,~\cite{clip,miech2019end2end,wray2019fine} observed that object state and action recognition implicitly emerges, to some extent, from these vision and language models.
In fact, the aligned vision and text training data often provides detailed descriptions of actions, objects and their different states.
In contrast to these works, we explicitly model the causal nature of actions and their impact on object states in order to leverage this strong inductive bias in our model. % than vision-language alignment alone.

\paragraphcustom{Object attributes and action modifiers.}
Learning object attributes (\eg \emph{sliced}, \emph{diced}) has been approached in a supervised manner in still images~\cite{misra2017red,Naeem2021GraphEmbeddings,Nagarajan2018AttributesOperators,Purushwalkam2019TaskDrivenNets,Yang2020UnseenConcepts} with the focus on the compositional nature of the attributes. % that act as modifiers for the different types of objects. %
Similarly, others have studied learning modifiers of actions (\eg \emph{quickly})~\cite{Doughty2020ActionModifiers} from short clips (20 seconds) mined from web-instructional videos.
Related to this, Doughty \etal~\cite{doughty2018who} analyzed how the visual changes of object states can be used for skill determination in videos.
The compositionality of natural language has also been explored for learning factored video-language embeddings for actions, objects and their attributes for retrieval applications~\cite{wray2019fine} or to learn a contextualized language-object embedding~\cite{COBE}.
%, but without explicit modelling of action-object interactions.  
%
Explicit models of changes in object states and the associated state-modifying actions have been explored in egocentric videos~\cite{FathiActionStateChange,Liu2017Fluents}. 
Others have considered significantly reducing the amount of supervision by learning object states from web images gathered by querying a web-image search engine~\cite{Isola2015DiscoveringStates}. % and we build on this work. 
Closely related to us, there is work on unsupervised learning of object states and the state-modifying actions from video~\cite{Alayrac16ObjectStates, dima2014youdo}. However, their focus is on small-scale learning from a set of trimmed and curated videos~\cite{Alayrac16ObjectStates} or a constrained scenario of videos observing a single specific scene~\cite{dima2014youdo}. 
In contrast, we consider large-scale learning from noisy untrimmed videos from the~web. %and develop an approach that can handle such ``in the wild" set-up. 

\paragraphcustom{Ordering as a form of supervision.}
The arrow of time is a strong signal~\cite{wei2018learning} to learn about actions.
Indeed, many actions happen in a certain order~\cite{fernando15modeling}. For example, you need to open a bottle before being able to pour something from it. This can be used as a source of supervision.
Past work~\cite{alayrac16unsupervised,bojanowski14weakly,bojanowski15weakly,Kukleva_2019_CVPR,Sener_2018_CVPR,Wang_Transformation,Zhang_2020_CVPR} has leveraged the supervision contained in such ordering to discover and temporally localize actions in untrimmed videos.
Similarly, the natural ordering of recurring events has been used to distinguish key events from the background~\cite{Zhukov2020Actionness}.
\cite{zhou2016learning} trained a generative model from time-lapse videos to generate the future state of an object altered by time.
Others have looked at the related task of next frame or action prediction as another form of supervision~\cite{Chang2020ProcedurePlanning,Girdhar2021AVT,luc2020transformation,mathieu2015deep,miech2019leveraging,srivastava2015unsupervised,ranzato2014video}.
In contrast, we use as the supervisory signal the strong causal ordering constraints that relate states of objects and the state-modifying actions.

\paragraphcustom{Action recognition and localization.}
The problem of detecting, classifying and localizing human actions has been extensively addressed by methods exploring motion and temporal evolution of appearance in a video.
Models for action recognition typically operate on short video clips trimmed to encompass a single action. Such models employ a mix of 2D and 3D convolutions~\cite{wang16temporal,carreira2017quovadis,Feichtenhofer_2019_ICCV,Feichtenhofer_2020_CVPR} or transformers and temporal attention~\cite{Fan_2021_ICCV,bertasius2021space,yan2022multiview}. Action localization methods often generate action proposals in the temporal domain using special modules such as graph neural networks~\cite{bai2020boundary,Buch_2017_CVPR,Lin_2019_ICCV,Xu_2020_CVPR,nawhal2021activity}. However, such methods typically require video annotations in terms of temporal action boundaries for training. 
Our proposed methods does not require temporal supervision. It uses changes in object states as a guidance for action localization.

\paragraphcustom{Object and action video datasets.} 
Most existing video action recognition datasets primarily contain state-preserving actions such as \textit{dancing} or \textit{playing a flute} \cite{carreira2017quovadis,kuehne2011hmdb,soomro2012}.
EPIC-KITCHENS~\cite{damen2018scaling}, Breakfast~\cite{kuehne2014language}, CrossTask~\cite{Zhukov2019Crosstask} or COIN~\cite{Tang2019COIN} datasets provide action sequence and object annotations for each video, but do not provide annotations related to changes of object states. %state-changing information.
HowTo100M~\cite{miech19howto100m}, YouCook2~\cite{youcook2} and RareAct~\cite{miech20rareact} datasets contain videos with state-changing actions; %, possibly even with free-form temporally localized captions~\cite{miech19howto100m,youcook2}.
however, they also do not provide clearly defined object-state annotations. % and the related state-modifying actions.
Closely related to us, Alayrac \etal\cite{Alayrac16ObjectStates} introduced an annotated video dataset of state-changing actions. However, this dataset was carefully curated to ensure that each video contains the action and object state change of interest.
Consequently, their dataset is small scale and contains only seven object-action classes with only tens of videos for each class. % compared to our dataset.
The Task-Fluent dataset~\cite{Liu2017Fluents} features several state-changing actions but is restricted to only 809 ego-centric videos.
In contrast, our dataset is 54$\times$ and 42$\times$ larger than the datasets of ~\cite{Alayrac16ObjectStates} and~\cite{Liu2017Fluents}, respectively, and contains untrimmed videos of a large variety of 44 different object-action classes.  
Concurrent to our work, a recently collected EGO4D~\cite{Grauman2021Ego4D} dataset contains 3,025 hours of ego-centric videos and also provides state change and action annotations. Our dataset is of comparable size but has a different complementary focus on untrimmed videos from the web.

%%%%%%%%%%%%%%%%%%%%%%

\section{Learning of actions and object states from untrimmed web videos}\label{sec:method}

We are given a set of web videos $v\in\mathcal{V}$ of arbitrarily length likely to depict a common state-modifying action applied on an object.
For example $\mathcal{V}$ can be a collection of birthday celebration videos, which are all likely to contain people \emph{blowing out candles} (\ie action) and changing the state of the candles from \emph{lighted} (\ie initial state) to \emph{extinguished} (\ie final state).

Given this, our goal is twofold: 
\textbf{(i)}  learning an action classifier $g$ that can recognize the action of interest and
\textbf{(ii)} learning a state classifier $h$ categorizing the modified object into an \emph{initial state} and an \emph{end state}.
We seek to do so without access to ground truth labels for the actions nor the object states. 
Instead, we design an approach that relies on the supervision provided by the causality of time: the action should appear between the two object states. 
In addition, we show that a handful of additional labelled exemplar images depicting the two object states help to make our approach significantly more robust to the noise in the training data via our new noise adaptive learning objective. 

In detail, the proposed learning procedure, illustrated in Figure~\ref{fig:overview}, optimizes the following objective:
\begin{equation}
	\min_{g,h} \sum_{v \in \mathcal{V}} \mathcal{L}_{(g, h)}(v, l(v))  \quad \text{(Sec. \ref{sec:model_optimization})},
	\label{eq:main_objective}
\end{equation}
where $h$ and $g$ are the learnt state and action classifiers, respectively, $\mathcal{L}$  is a loss function adapted to the noisy nature of web videos and
$l(v)$ are labels for both the action and state temporal positions within the video $v$.   
Since these labels are not given in advance, we instead estimate it on the fly during the optimization procedure via the following maximization:
\begin{equation}
	 l(v) = \argmax_{l \in \mathcal{D}_v} S_{(g,h)}(v, l) \quad \text{(Sec. \ref{sec:model_constrains})},
	 \label{eq:main_inference}
\end{equation}
where $S$ is a scoring function that depends on the current action and state classifiers $g$ and $h$.
$\mathcal{D}_v$ is the set of labels that respect our temporal causality constraints. The learning proceeds by iteratively \textbf{(i)} learning action and state classifiers, $g$ and $h$, given the current labels $l$ of the input videos (Eq.~\eqref{eq:main_objective} and Sec.~\ref{sec:model_optimization}) and \textbf{(ii)} finding the labels~$l$ of the videos that respects the causal ordering constraints given the current action and state classifiers, $g$ and $h$ (Eq.~\eqref{eq:main_inference} and Sec.~\ref{sec:model_constrains}). 
Details about these two steps are provided next.

% Inference step / estimation step
\subsection{Noise adaptive learning objective}\label{sec:model_optimization}

In this section, we describe the loss function $\mathcal{L}$ from Equation~\eqref{eq:main_objective}.
Each video $v$ is represented by a sequence of $T_v$ $d$-dimensional visual features: $v=\{x_t\}_{t=1}^{T_v}$. 
Each $x_t\in\mathbb{R}^d$ describes a temporal segment one second long of the original video. 
Here, we assume that the labels $l(v)$ are known for all videos, \ie the temporal position of the action $l_a(v) \in  \llbracket1,T_v \rrbracket$ as well as the temporal positions of the initial state $l_{s_1}(v)\in \llbracket1,T_v\rrbracket$ and the end state $l_{s_2}(v)\in \llbracket1,T_v\rrbracket$ (see Section~\ref{sec:model_constrains} for details about how $l(v)$ is obtained).

\paragraphcustom{Action and state classifiers.}
The goal here is to learn the action and state classifiers, $g$ and $h$, given the labels $l$.
The action classifier $g$ takes as input a visual feature $x\in\mathbb{R}^d$ and outputs a confidence score $g(x)\in[0, 1]$ that the feature depicts the action of interest.
Similarly, the state classifier $h$ takes as input the visual feature $x$ and outputs two scores $h_1(x),\ h_2(x)\in[0, 1]$ % and $h_2(x)\in[0,1]$ (with $h_1(x)+h_2(x)=1$)
giving an estimate of probability that the feature corresponds to the initial and the end state.

\paragraphcustom{Loss definition.}
Formally, the loss function $\mathcal{L}_{(g,h)}$ of a video $v$ and its associated labels $l(v)$ is a weighted sum of losses $\mathcal{L}_{g}$ for the action and $\mathcal{L}_{h}$ for the states:
\begin{equation}
    \mathcal{L}_{(g,h)}(v, l(v)) = \omega(v) \big(\mathcal{L}_h(v, l(v)) + \lambda\mathcal{L}_g(v, l(v))\big)
    \label{eq:loss}
\end{equation}
where $\lambda$ controls the relative importance of the two partial losses, and $g$ and $h$ are the action and state classifiers, respectively, that are being learnt. 
Given the noisy nature of the dataset of untrimmed videos obtained from the web, we weight each video's contribution to the overall loss function by a scalar weight $\omega(v)$, which is deduced from comparing the video frames to a small set of exemplar images  (Figure~\ref{fig:overview}, bottom left) and described later.

The action and state losses in Eq.~\eqref{eq:loss} are cross-entropy losses applied on the output of the classifiers as:
\begin{align}
    \mathcal{L}_g(v, l(v)) & = -\mu\sum_{t\in\mathcal{A}^P_v} \log g(x_t) - \sum_{t\in\mathcal{A}^N_v} \log\big(1 - g(x_t)\big)\nonumber\\
    \mathcal{L}_h(v, l(v)) & = -\sum_{t\in\mathcal{S}^1_v} \log h_1(x_t) -\sum_{t\in\mathcal{S}^2_v} \log h_2(v_t)
\end{align}
where $\mathcal{S}^1_v$, $\mathcal{S}^2_v$, $\mathcal{A}^P_v$ are sets of positive examples deduced from $l(v)$ where the model is expected to predict the initial state, the end state and the action, respectively. 
The set $\mathcal{A}^N_v$ contains negative examples where the model is expected to produce the \textit{no-action} label.
We describe how these sets are deduced from the current labels $l(v)$ of the video below.
The parameter $\mu$ is the relative weight between the action/no-action class.

\paragraphcustom{Sampling of positive examples.}
All the positive sets $\mathcal{S}^1_v$, $\mathcal{S}^2_v$, $\mathcal{A}^P_v$ are sampled in the same way and are directly obtained from labels $l(v)$.
They all contain feature indices within a temporal window centered on the currently estimated locations of the initial state $l_{s_1}(v)$, end state $l_{s_2}(v)$ and the action $l_a(v)$.
Formally, the set of positive examples for the initial state $t\in\mathcal{S}_v^1$ is defined as:
\begin{equation}
    \mathcal{S}_v^1 = \big\{ t: |t-l_{s_1}(v)|\leq \delta,\ 1\leq t\leq T_v\big\}
\end{equation}
where $l_{s_1}(v)$ is the currently estimated position of the initial state in video $v$, $T_v$ is the length of the video, and $\delta$ is a hyper-parameter defining the number of the neighbouring features considered as positive. The intuition is that we wish to consider as positives several temporally nearby examples (within the temporal window defined by $\delta$) as they are likely to also contain the initial object state.  
The sets of positive examples for the end state and action, $\mathcal{S}^2_v$ and $\mathcal{A}^P_v$, are defined analogously.

\paragraphcustom{Sampling of \textit{no-action} examples.} 
There are various ways of sampling the set $\mathcal{A}^N_v$ of \textit{no-action} examples. 
Considering all negatives in the video, $\mathcal{A}^N_v=\{t:t\notin \mathcal{A}^P_v\}$, is impractical due to class imbalance which is a) directly dependent on the length of the video and b) extremely large with ratios that can exceed 1 to 100 in long videos. 
Instead, we opt for defining $\mathcal{A}^N_v$ as a set of video feature indices at a given distance $\kappa$ from the location $t'\in\mathcal{A}^P_v$ of the positive example:
\begin{equation}
    \mathcal{A}^N_v = \big\{ t: t'\in \mathcal{A}^P_v,\ |t-t'| = \kappa,\ 1\leq t\leq T_v\big\}.
\end{equation}
The intuition is that, for appropriate $\kappa$, set $\mathcal{A}^N_v$ will contain {\em hard} negatives, that are visually related to the correct action but yet negative. If $\kappa$ is too small, $\mathcal{A}^N_v$ will contain positive examples, which will harm training the action classifier. On the other hand, if $\kappa$ is too large,  $\mathcal{A}^N_v$  can contain unrelated (easy to discriminate) actions from the rest of the video. In Section \ref{sec:experiments}, we ablate the choice of $\kappa$ and show that this definition of negatives for the action classifier is crucial for obtaining good performance compared to randomly sampling the positions of the negatives.
Lastly, we note that there can be positions in a video that are not in any of the four $\mathcal{S}^1_v$, $\mathcal{S}^2_v$, $\mathcal{A}^P_v$, $\mathcal{A}^N_v$ sets. 
Actually, in the case of longer videos, most of the segments are without any label and thus do not contribute to the loss.

\paragraphcustom{Noise adaptive weighting from a few exemplar images.} 
As our training videos are obtained automatically from the web without any manual curation, a large proportion of videos may contain unrelated content and thus harm the performance of the model. To address this issue, we download a small number of images (up to five) via Google Image search containing the object of interest in both initial and end states. Then we use a pre-trained visual model applied in a zero-shot manner together with the causal ordering constraint to compute video relevance score $r_v$ as follows:
\begin{equation}
    r_v=\max_{t<t'} \sum_{e_1\in\mathcal{E}_1}\textrm{sim}\left(e_1,v_t\right) \sum_{e_2\in\mathcal{E}_2}\textrm{sim}\left(e_2,v_{t'}\right)
    \label{eq:video_scores}
\end{equation}
where $\mathcal{E}_1$, $\mathcal{E}_2$ are sets of exemplar images representing the initial and the end state, respectively, and $\textrm{sim}\left(e,v_t\right)$ is the similarity between the exemplar image $e$ and the video feature at the $t$-th temporal location of video $v$ computed as cosine similarity of features extracted by a pre-trained visual model. We use this relevance score to weigh the contribution of each video in the loss function using the following weight:
\begin{equation}
    \omega(v) = \sigma\left(\frac{r_v - \theta}{\tau}\right)
    \label{eq:relevance_weight}
\end{equation}
where $\sigma$ is sigmoid function, $\tau$ is a temperature hyper-parameter and $\theta$ is a centering hyper-parameter. The relevance weight $\omega(v)$ varies between 0 and 1. The weight is close to 0 for videos that do not have any frames similar to the object state exemplar images $e$ satisfying the causal ordering constraint. On the other hand, the weight is close to 1 for videos, which have frames with high similarity to exemplar images $e$ and which satisfy the causal ordering constraint.  
%Otherwise, the weight is close to 1.
As a result, this weight is effective in suppressing irrelevant videos during the learning process.
Note that we do not use exemplar action images as we found their collection at scale to be problematic.
%{Note we do not have exemplar images for actions as we observed it is much more difficult to collect them.}
%
We select $\theta$ for each object-action category independently because the relevance scores $r_v$ vary greatly depending on the exemplar images or the video content. We use $\theta$ that minimizes intra-class variance of the relevance scores. More details are given in \ARXIVversion{Appendix~\ref{supsec:omega}}{the appendix~\cite{soucek22lookforthechange}}.

%\subsection{Finding labels via causal ordering constraints}\label{sec:model_constrains}
\subsection{Labelling with causal ordering constraints}\label{sec:model_constrains}
In this section, we explain how labels $l(v)$ identifying the best action and object state locations in video $v$ are automatically obtained given the current, possibly sub-optimal, action and state classifiers, $g$ and $h$. 
More formally, we assume we are given fixed classifiers $g$ for an appearance-changing action and $h$ for the manipulated object's states. We are also given a video $v$ containing the action exerted upon the object with high probability. Then to compute the most likely location of the action $l_a(v)$, the initial state $l_{s_1}(v)$ and the end state $l_{s_2}(v)$, 
we employ predictions of the current action $g$ and object state classifiers $h$ as follows:
\begin{align}
	l(v) = \argmax_{l\in\mathcal{D}_v} \ & h_1(x_{l_{s_1}})\cdot g(x_{l_a}) \cdot h_2(x_{l_{s_2}}) %\lambda \log g(v_j) + \mathcal{A}(i,j,k)
	\label{eq:argmax}
\end{align}
where $\mathcal{D}_v$ is a set of all possible locations of the action and the object states satisfying the {\em causal ordering constraint}, $h_1(x_{l_{s_1}})$ is the output of the initial state classifier $h_1$ at temporal location $l_{s_1}$ in video $v$, $h_2(x_{l_{s_2}})$ is the output of the end state classifier $h_2$ at temporal location $l_{s_2}$, and $g(x_{l_a})$ is the output of action classifier $g$ at temporal location $l_{a}$. In other words, the goal is to identify object state and action locations in the video that satisfy the causal ordering constraint and maximize the product of output scores of the state and action classifiers as given in Eq.~\eqref{eq:argmax}. 

\paragraphcustom{Causal ordering constraint.} 
The causal ordering constraint employed in Eq.~\eqref{eq:argmax} is motivated by the fact that many object-modifying actions cannot be physically reversed, \eg \textit{cut apple}. Also, many object-modifying actions are commonly performed only one way, even if the other direction is physically possible, \eg \textit{clean shoes}. Thus we restrict the set of permissible locations of actions and states $\mathcal{D}_v$ to follow the order of initial object state $\rightarrow$ manipulating action $\rightarrow$ end object state. Formally, we define the set $\mathcal{D}_v$ of  labels satisfying this constraint as 
\begin{equation}
	\mathcal{D}_v=\big\{l: 1\leq l_{s_1}<l_a< l_{s_2} \leq T_v \big\}
\end{equation}
where $T_v$ is the length of the video $v$ and $l_{s_1}$, $l_{a}$, $l_{s_2}$ are the temporal positions of the initial state, action and the end state, respectively.
In untrimmed videos from the web, it is common to have multiple instances of objects of interest or distracting objects in the same video. The same is true for actions. The ordering constraint pinpoints the most prominent instance of the object and action in each video. Other instances are ignored and treated as background.

\section{\datasetname{}: a state-changing actions dataset}\label{sec:dataset}
%\section{\datasetname{}: a dataset of state-changing actions}\label{sec:dataset}
Our goal is to automatically learn the different states of objects together with the state-modifying actions without the need for manual curation of videos.
To this end, we collect a new large-scale dataset of more than 34,000 in-the-wild untrimmed videos (more than 2600 hours of video) covering 44 various state-changing actions.
Our state-changing actions depict a wide range of human activities such as: \textit{cleaning shoes} (initial state: dirty shoes, action: cleaning, final state: clean shoes), \textit{cut avocado} (initial state: entire avocado, action: cutting, final state: cut avocado) or \textit{gift wrapping} (initial state: unwrapped gift, action: wrapping, final state: wrapped gift).
We emphasize the training split of our \datasetname{} dataset is collected in such a way that only a name of an action is needed to be specified. No other manual selection, acquisition or annotation is required.
Next we describe our dataset collection process and provide detailed statistics of \datasetname{}.

%\subsection{Dataset collection}
\paragraphcustom{Dataset collection.}
First, we select a set of state-changing actions.
Such actions imply a modification of the appearance of an object through manipulation.
We restrict ourselves mostly to irreversible actions in order to eliminate scenarios where two actions manipulate an object from an initial state back to the same initial state via an end state, such as \textit{open} and \textit{close a door}.
However, we allow actions such as \textit{clean shoes} as these are not considered immediately reversible because we do not expect the shoes to be immediately dirty after being cleaned.
We also require the actions to be sufficiently represented on YouTube.

Given these conditions, we devise a set of 44 state-changing actions. % \hideme{Please see Table~\ref{tab:per-class} for the full list.}
Note that some of the classes contain similar or complementary actions such as \textit{cut} and \textit{peel an avocado}.
For each action, we query YouTube with queries such as \textit{``How to clean shoes?''} and download up to two thousand retrieved videos. We exclude videos longer than 15 minutes and obtain 34,428 videos with a total duration of 2,642 hours and an average video duration of 4.6 minutes. 
We have manually annotated a small fraction of videos for evaluation with the following labels: \textit{background}, \textit{initial state}, \textit{action}, \textit{end state}. In total we have annotated 667 videos summing up to 48 video hours and yielding 15 video samples per state-changing action on average.
Please see \ARXIVversion{Appendix~\ref{supsec:ds}}{the appendix~\cite{soucek22lookforthechange}} for more details on the action selection, annotation, and additional dataset statistics.

%%%%%%%%%%%%%%%%%%%%%%%%%%%%%%%%%%%%%%%%%%%%%
\section{Experiments}\label{sec:experiments}
%%%%%%%%%%%%%%%%%%%%%%%%%%%%%%%%%%%%%%%%%%%%%

\begin{figure*}
  \centering
  \begin{subfigure}{0.39\linewidth}
    \centering
    \includegraphics[height=3.98cm]{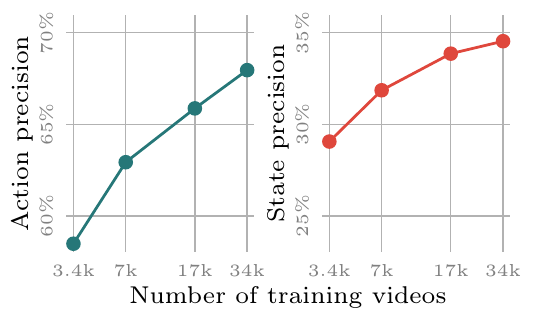}
    \vspace*{-1mm}
    \caption{\textbf{Dataset size}}
    \label{fig:ds_size}
  \end{subfigure}
  \hfill
  \begin{subfigure}{0.33\linewidth}
    \centering
    \includegraphics[height=3.98cm]{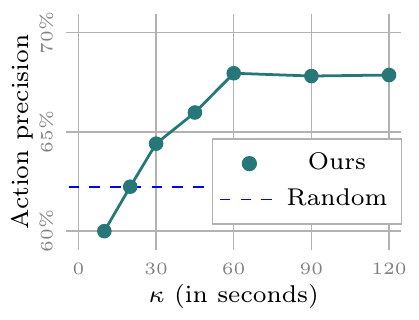}
    \vspace*{-1mm}
    \caption{\textbf{Negative sampling}}
    \label{fig:ds_kappa}
  \end{subfigure}
  \hfill
  \begin{subtable}{0.27\linewidth}
  \centering
  {\footnotesize
  \vspace*{1.3cm}
  \begin{tabular}{l@{~}c@{~}c@{~}}
    \toprule
    Method & St prec. & Ac prec. \\
    \midrule
    \textbf{a.} Ours                                       & \textbf{0.35} & \textbf{0.68} \\
    \textbf{b.} w/o noise adapt.                          & 0.34 & 0.64 \\
    \textbf{c.} w/o multi-layer                            & 0.34 & 0.61 \\
    \textbf{d.} w/o data augment.                          & 0.33 & 0.66 \\
    \bottomrule
  \end{tabular}
  \vspace*{0.6cm}
  }
  \vspace*{-1mm}
  \caption{\textbf{Selected variants of our model}}%Selected ablated variants of our model.}
  \label{tab:osads_ablations}
  \end{subtable}
  \vspace*{-3mm}
  \caption{{\bf Ablations on the \datasetname{} dataset.} (a) State and action precision with the increasing size of the training dataset. (b) Action precision as a function of the distance $\kappa$ of sampled negatives from the positives compared with random negative sampling (Random). (c) Ablations of different components of our model.}
  \label{fig:all_ablations}
  \vspace*{-5mm}
\end{figure*}

In this section, we first describe the model architecture and training regime (Section~\ref{subsec:impl_details}). % and data pre-processing and report all hyper-parameter values (Section~\ref{subsec:impl_details}).
Then we describe the datasets we test on, the evaluation process, and the metrics we report (Section~\ref{subsec:data_metrics}). We ablate our most important design decisions and show their benefit in Section~\ref{subsec:ablations}. Finally, we compare our method with related work and various baselines and show qualitative results in Section~\ref{subsec:visualizations}.

\subsection{Implementation details}\label{subsec:impl_details}
\noindent\textbf{Architecture.} Our definition of the model learning procedure (Section~\ref{sec:method}) allows us to use any differentiable temporal video classifier for $g$ and $h$.
We choose both classifiers to be applied on the same visual features and share the same architecture: a two-layer MLP with hidden dimension of 512 and ReLU activation function. 
We train separate classifiers for every dataset category, thus the action classifier outputs a single scalar followed by the sigmoid activation function, the state classifier outputs two scalar followed by the softmax activation function. 
The feature extractors operate on original videos and downsample the temporal resolution of the video features $x_t$ into one frame per second. 
For the feature extractors, we use 2D ResNeXT pre-trained on ImageNet-21K \cite{OurResnextWeights} and 3D TSM ResNet50 pre-trained on HowTo100M and AudioSet \cite{Alayrac2020MMVN}. 
The 2D and 3D features are concatenated prior to be fed to the classifiers.
We initialize the MLPs randomly~\cite{He2015initializer}. 
We do not back-propagate gradients into the feature extractors.

\paragraphcustom{Training regime.} During training, we sample a batch of videos and compute action and states locations for each video in the batch (Eq. \eqref{eq:argmax}). Then we compute gradients of the loss function $\mathcal{L}_{(g,h)}(v)$ with respect to the model parameters and perform one step of gradient descent with a momentum. We alternate these steps for 100 epochs. Additional details with all hyper-parameter values and data preprocessing steps are in \ARXIVversion{Appendix~\ref{supsec:trainDetails}}{the appendix~\cite{soucek22lookforthechange}}.

%\subsection{Datasets and evaluation metrics}\label{subsec:data_metrics}
\subsection{Evaluation protocol and metrics}\label{subsec:data_metrics}
We evaluate on both our new \datasetname{} dataset and on the dataset of Alayrac~\etal~\cite{Alayrac16ObjectStates}. The first one depicting noisy untrimmed videos and the latter one depicting short curated trimmed videos.
For each video we predict the location of the action and the initial and end states by Equation~\eqref{eq:argmax} using predictions of the trained classifiers. For Alayrac \etal dataset, we average predictions for frames corresponding to a single so-called \textit{tracklet} and apply Equation~\eqref{eq:argmax} on the averaged predictions only. 
In all experiments including ablations, we report performance of the best performing model weights from training averaged over three runs.

We follow the related work \cite{Alayrac16ObjectStates} with metrics and report precision for both the action and the states. 
For a given video, action precision is either one or zero depending on whether the predicted action location is within the ground truth interval. 
For state precision, a value of 0.5 is also possible if only one of the two state locations matches the ground truth. The video-level metrics are then averaged over all videos in a category, and finally averaged over all categories to suppress effects of differences in distributions of videos throughout the categories.

\subsection{Ablations}\label{subsec:ablations}
In this section, we ablate the key components of our model and show their benefit. We investigate the effect of dataset size, model depth and input data augmentation on performance. We also show the importance of sampling action negatives and noise adaptive video weighting during training. All experiments are performed on the full \datasetname{} dataset except for the dataset size ablation.

\paragraphcustom{Dataset size matters.} 
We train our model on various fractions of the dataset to investigate the effect of the dataset size on performance. 
In each category, we sort videos according to their relevance score $r_v$ (Eq.~\eqref{eq:video_scores}) and train our model with the top scoring 5\%, 10\%, 20\%, 50\% and 100\% of videos. Figure~\ref{fig:ds_size} shows the effect of varying the dataset size on action and state precision. There is a clear improvement with increasing dataset size. Also, we do not observe performance saturation indicating further improvements could be achieved with even larger datasets. 
This observation is even more interesting when we consider that many low scoring videos may not contain the action or even the object of interest, yet the model improves even in this low signal-to-noise ratio setting. 

\paragraphcustom{Sampling action negatives.} In our experiments, we show that the strategy for sampling action negatives $\mathcal{A}_N$ is an important design choice significantly affecting the model's action localization performance. When the negatives are sampled uniformly at random, as shown in Figure~\ref{fig:ds_kappa} in blue, we see a drop from 68\% to 62\% in action precision compared to our method of using negatives that are at a fixed distance $\kappa$ from the positives. We also test different values of $\kappa$ (Figure~\ref{fig:ds_kappa}). We can see the best performance is achieved using $\kappa \geq 60$. We hypothesize that by using smaller values of  $\kappa$ the action negatives can still contain the action which hinders model performance.
Nonetheless, most $\kappa$ values~still outperform the random sampling.
Increasing $\kappa$ further yields only a minor change in the precision as most of the videos are 1-5 minutes long and thus the negatives often lay outside of the video. In such cases, the first or the last frames of the video are taken as negatives.

\paragraphcustom{Noise adaptive weighting.} 
We test the added benefit of using video relevance scores to weigh individual videos in the batch. These relevance scores incorporate the only truly manual data collection in the whole method, even if very minor as only a handful of images are required. 
Results without the noise adaptive video weighting are shown in Figure~\ref{tab:osads_ablations} (row \textbf{b. w/o noise adapt}) and show a clear drop in performance, especially for actions compared to the full method (row \textbf{a. Ours}).  

\paragraphcustom{Other ablations.} We test the benefits of our two-layer MLP on top of a feature extractor compared to a single linear layer used in related work~\cite{Alayrac16ObjectStates}. 
In Figure \ref{tab:osads_ablations}, we see a clear benefit of using the two-layer MLP \textbf{a. Ours} compared to a linear classifier \textbf{c. w/o multi-layer} in the same set-up suggesting that a linear classifier used by Alayrac \etal \cite{Alayrac16ObjectStates} for it's closed-form solution is insufficient to discriminate between states and actions.
We also add augmentation for the input videos and show the benefit of this standard practice in neural network training in our setup (Figure~\ref{tab:osads_ablations}, row~\textbf{d. w/o data augment.}). Additionally, in \ARXIVversion{Appendix~\ref{supsec:featAblation}}{the appendix~\cite{soucek22lookforthechange}}, we show the effect of different feature extractor backbones on the final performance.

\subsection{Comparison with the state-of-the-art}\label{subsec:comparison}

\noindent\textbf{Compared methods.} 
We compare our method to several strong baselines described next.
\textbf{(a) Alayrac \etal~\cite{Alayrac16ObjectStates}.} We compare results with the state-of-the-art approach for learning object states and actions from video by Alayrac \etal~\cite{Alayrac16ObjectStates}, which  learns a linear classifier on fixed video features using discriminative clustering~\cite{bach07diffrac}. 
 As we notice this method could be unstable, we report the best numbers reached during the course of optimization.
\textbf{(b) CLIP~\cite{clip}.} We compare to the zero-shot CLIP approach~\cite{clip} that has demonstrated strong results on a large variety of recognition tasks and thus presents a strong baseline. We obtain the state and action classifiers by projecting the textual descriptions of our action and object state classes into the joint text-image space. % where similarities to the descriptions is measured.
We employ prompt engineering by producing multiple text descriptions for each state and action and report the performance of the best one. To make the comparison as fair as possible, we employ the causal ordering constraint for computing the state and action precision as in Equation \eqref{eq:argmax}. 
\textbf{(c) MIL-NCE S3D~\cite{miech2019end2end}} Analogously to CLIP, we also compare to the zero-shot video-based S3D model trained on the HowTo100M dataset~\cite{miech19howto100m} using MIL-NCE loss \cite{miech2019end2end}. We use the same evaluation procedure as for CLIP with the same text descriptions and the causal ordering constraint employed.
\textbf{(d) Image examples.} We use the images gathered for the noise adaptive weighting and measure their similarity to individual video frames in a feature space. 
We use the causal ordering constraint for the computation of the state precision metric. 
\textbf{(e) Random.} We also report chance performance with the state ordering constraint employed.

\paragraphcustom{Comparison on the dataset of Alayrac \etal.}
In Table~\ref{tab:iccv17ds_results} we report performance on the dataset from Alayrac~\etal~\cite{Alayrac16ObjectStates} containing curated trimmed videos of seven action-object classes. 
Note that some dataset classes contain only tens of videos thus we use additional temporal attention using $sim(\mathcal{E}_*,v_t)\cdot h_*(x_t)$ instead of $h_*(x_t)$ in Equation \ref{eq:argmax}. 
Without it, our method can focus on different temporally consistent actions that are also present in the same video, \eg \textit{jack up a car} instead of \textit{remove a wheel}. The results in Table~\ref{tab:iccv17ds_results} demonstrate the benefits of our approach over~\cite{Alayrac16ObjectStates}, despite fact that~\cite{Alayrac16ObjectStates} uses much stronger supervision in the form of a pre-trained object detector for each objects.

\begin{table}
  \centering
  {\small
  \begin{tabular}{lcc}
    \toprule
    Method & St prec. & Ac prec. \\
    \midrule
    Random & 0.09 & 0.22 \\
    CLIP ViT-L/14 \cite{clip} & 0.42 & 0.42 \\
    Alayrac \etal \cite{Alayrac16ObjectStates}\textsuperscript{$\dagger$} & 0.48 & 0.55 \\
    Ours  & \textbf{0.49} & \textbf{0.58} \\
    \bottomrule
    \multicolumn{3}{l}{\footnotesize \textsuperscript{$\dagger$} Trained with known object bounding boxes.}
  \end{tabular}
  }
  \vspace*{-3mm}
  \caption{Comparison to the state-of-the-art approach~\cite{Alayrac16ObjectStates} on their own dataset.
  }
  \vspace*{-3mm}
  \label{tab:iccv17ds_results}
\end{table}

\paragraphcustom{Results on our new \datasetname{} dataset.} In Table \ref{tab:osads_results} we report quantitative results on our much larger \datasetname{} dataset that contains long untrimmed uncurated videos. 
We observe that the zero-shot CLIP method \textbf{(b)} and the image-based approach \textbf{(c)} match the state-of-the-art approach of \textbf{(a)} Alayrac \etal \cite{Alayrac16ObjectStates}.
Our method produces significantly better results and outperforms the state-of-the-art approach of Alayrac \etal \cite{Alayrac16ObjectStates} as well as the other baselines, demonstrating the benefits of our approach on noisy untrimmed videos.
The full set of per-class results on the entire set of 44 classes is in \ARXIVversion{Appendix~\ref{supsec:results}}{the appendix~\cite{soucek22lookforthechange}}. 

\begin{table}
  \centering
  {\small
  \begin{tabular}{c@{~}lcc}
    \toprule
    \multicolumn{2}{l}{Method} & St prec. & Ac prec. \\
    \midrule
    \textbf{(e)}  & Random w/ constraint                          & 0.15 & 0.41 \\
    \textbf{(d)} & Image examples                                & 0.29 & -    \\
    \textbf{(c)}& MIL-NCE S3D \cite{miech2019end2end}& 0.27 & 0.50 \\
    \textbf{(b)} & CLIP ViT-L/14 \cite{clip} & 0.30 & 0.63 \\
    \textbf{(a)}  & Alayrac \etal \cite{Alayrac16ObjectStates}    & 0.30 & 0.59 \\
    & Ours                                       & \textbf{0.35} & \textbf{0.68} \\
    \bottomrule
  \end{tabular}
  }
  \vspace*{-3mm}
  \caption{Comparison to state-of-the-art on our \datasetname{} dataset.}
  %\caption{Comparison to state-of-the-art on our new \datasetname{} dataset.}
  \label{tab:osads_results}
  \vspace*{-5mm}
\end{table}

\begin{figure}[t]
    \centering
    \includegraphics[width=0.99\linewidth]{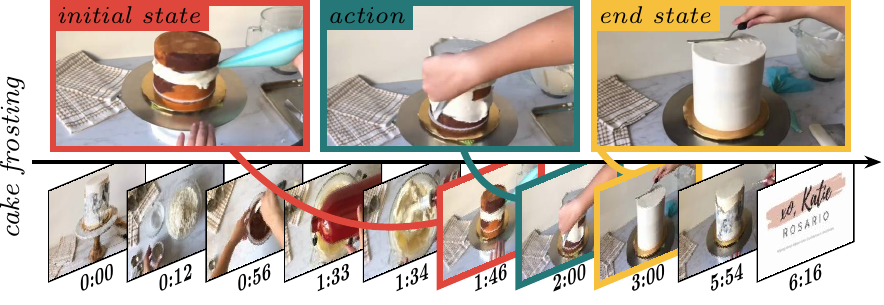}\\
    \vspace{1mm}
    \includegraphics[width=0.99\linewidth]{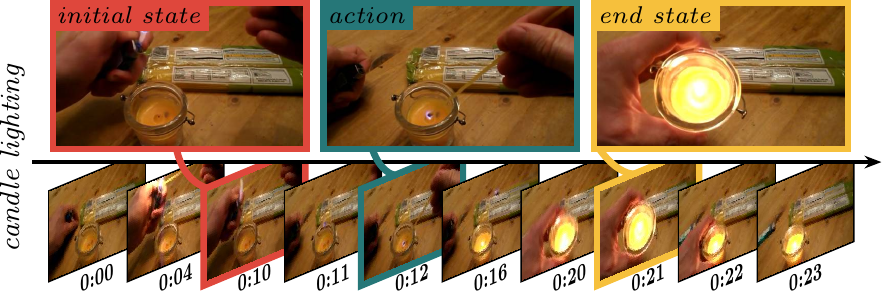}\\
    \vspace{1mm}
    % \includegraphics[width=0.99\linewidth]{final_teasers/plane_fhAbLwaymDU.pdf}\\
    % \vspace{1mm}
    \includegraphics[width=0.99\linewidth]{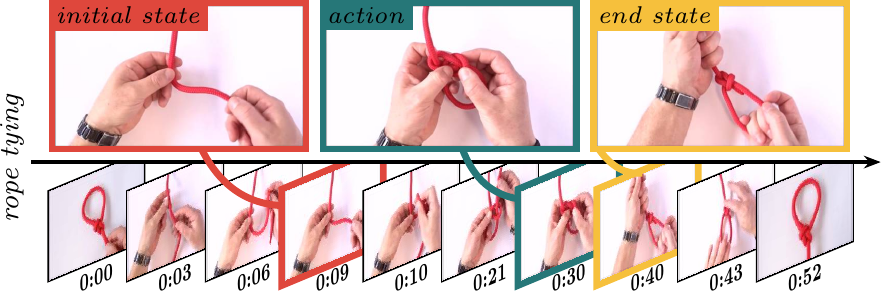}
    %\vspace*{-3mm}
    \vspace{-4pt}
    \caption{\small \textbf{Predicted object-state and action frames} (top 3 frames in each example) and their temporal localization on the video timeline (bottom). Note how the object states and the state-modifying actions are temporally localized in long uncurated videos from the web. See additional examples in \ARXIVversion{the appendix}{the appendix~\cite{soucek22lookforthechange}}.} 
    \label{fig:qual_examples}
    \vspace*{-5mm}
\end{figure}

%\subsection{Qualitative results}\label{subsec:visualizations}
\paragraphcustom{Qualitative results.}\label{subsec:visualizations}
Qualitative results are shown in Figures~\ref{fig:teaser} and~\ref{fig:qual_examples}\hideme{, and~\ref{fig:t-shirt_example}}. 
They demonstrate the ability of our approach to learn object state and action classifiers from long uncurated videos from the web. Additional qualitative results for a range of object-action classes are in \ARXIVversion{Appendix~\ref{supsec:results}}{the appendix~\cite{soucek22lookforthechange}}. 

\paragraphcustom{Limitations and societal impact.}
Our approach has currently three main limitations. 
First, it relies on training videos being available on the web. This may not be the case for all state-modifying actions. For example, some common or uninteresting daily actions (such as \textit{open a fridge}) may not be frequently captured and uploaded to Youtube.  
Second, in some cases the appearance variation of the object states is too large, \eg \textit{remove weed}, to be learnt in a fully unsupervised manner. 
Finally, currently we learn a separate model for each class from a set of videos for that class. Enabling sharing information during learning among models of related classes, \eg \textit{cut apple} and \textit{cut avocado}, remains an interesting open problem. Further discussion about potential negative societal impact of our work is provided in \ARXIVversion{Appendix~\ref{supsec:impact}}{the appendix~\cite{soucek22lookforthechange}}.

\section{Conclusion}\label{sec:conclusion}
We have developed a new approach for learning object-states and state-changing actions from noisy untrimmed videos from the web. Our approach relies on a novel noise-adaptive learning objective supervised by exemplar images together with causal ordering constraints temporally relating object changes and actions in videos. We have validated our approach on an existing dataset of state-modifying actions as well as our newly collected dataset of more than 2600 hours of videos from the web, demonstrating significant improvements compared to the state-of-the-art. This work opens-up the possibility of large-scale automatic learning of causal object-action relations for embodied video understanding and robotics.  

\paragraphcustom{Acknowledgements.}
This work was partly supported by the European Regional Development Fund under the project IMPACT (reg. no. CZ.02.1.01/0.0/0.0/15\_003/0000468), the Ministry of Education, Youth and Sports of the Czech Republic through the e-INFRA CZ (ID:90140), the French government under management of Agence Nationale de la Recherche as part of the ``Investissements d'avenir'' program, reference ANR19-P3IA-0001 (PRAIRIE 3IA Institute), and Louis Vuitton ENS Chair on Artificial Intelligence. We would like to also thank Kate\v{r}ina Sou\v{c}kov\'{a} and Luk\'{a}\v{s} Ko\v{r}\'{i}nek for their help with the dataset.

% 

%%%%%%%%% REFERENCES
{\small
\bibliographystyle{ieee_fullname}
\bibliography{egbib}
\flushcolsend
}

\ARXIVversion{\appendix

\ARXIVversion{\section*{Appendix}}{}
\ARXIVversion{We}{In this Supplementary Material, we} start by providing additional details on the selection of hyper-parameters for the video relevance weight $\omega(v)$ in Section \ref{supsec:omega}.
In Section \ref{supsec:trainDetails}, we describe training details, hyper-parameters, and data preprocessing steps. We then provide details on the action selection and annotation process for our new \datasetname{} dataset together with additional statistics in Section \ref{supsec:ds}. Section~\ref{supsec:featAblation} shows the effect of different feature extractors on the performance. Further, we report per-class quantitative results and show qualitative results in Section~\ref{supsec:results}. Lastly, we discuss broader impact of our work in Section~\ref{supsec:impact}. On the project website\footnote{\url{https://data.ciirc.cvut.cz/public/projects/2022LookForTheChange/}}, we also provide a video showing our model's predictions on handful of dataset videos.

\section{Video relevance weight $\omega(v)$}\label{supsec:omega}

Video relevance weight $\omega(v)$ \ARXIVversion{(Equation \ref{eq:relevance_weight})}{(Equation (8) in the main paper)} contains a temperature hyper-parameter $\tau$ and a centering hyper-parameter $\theta$.
We choose $\tau$ globally by grid search. However, $\theta$ needs to be chosen individually per each category as the distribution of relevance scores $r_v$ varies greatly between the categories. Possible value for $\theta$ would be a median or other fixed quantile of the score distribution. Instead, we opt for solution that does not require manual choice of a quantile and follows from our observation that the score distributions for different categories are often bimodal with the two modes corresponding to the relevant and irrelevant videos. We compute $\theta$ for each category $\mathcal{C}$ by minimizing the intra-class variance of the category relevance scores $r_v$~as:
\begin{equation}
    %\theta_\mathcal{C}= 
    \argmin_\theta \textrm{var}_{v\in\mathcal{C}}\{r_v: r_v < \theta\} + \textrm{var}_{v\in\mathcal{C}}\{r_v: r_v > \theta\}.
\end{equation}
We validate our approach by computing the number of annotated videos with $r_v>\theta$, as the annotated (test) videos contain the object of interest with certainty. Using this method, we retrieve 80.7\% of the annotated videos while retrieving only 59.5\% of all dataset videos. The fixed quantile method retrieves 77.8\% of the annotated videos while retrieving the same total number of videos.

\section{Training details}\label{supsec:trainDetails}
\noindent\textbf{Hyper-parameters.} We use a batch size of 48 randomly sampled videos. %
We optimize the classifiers using stochastic gradient descent with a momentum of $0.9$ and $L_2$ penalty of $0.001$. 
We sample five ($\delta=2$) positive examples for both the action and states and use temperature $\tau=0.001$ for the relevance weight $\omega(v)$. 
In order to compute the features used for the noise adaptive video relevance score $r_v$ we use the 2D ResNeXT backbone only. 
The distance parameter for action negatives is fixed to $\kappa=60$. 
The action loss is weighted by $\lambda=0.2$ and the action positives are weighed by $\mu=10$.

\paragraphcustom{Data preprocessing.} We apply data augmentation to the inputs as follows: each video is randomly rotated by up to five degrees and horizontally flipped with probability 50\%. Then each video's sides are randomly cropped by 16\% and with 80\% chance one change of brightness, color, or contrast is applied. 
The same augmentation is applied on all frames of the video to ensure temporal consistency.
The importance of data augmentation is shown in the ablation section of the main paper.
The visual features are extracted by running an image feature extractor on one frame per second and a video feature extractor on 25 frames per second of the original video. The output of the video extractor is temporally downsampled to match the 1fps sampling rate of the image features.

\section{ChangeIt dataset details}\label{supsec:ds}
%\section{ChangeIt dataset analysis}\label{supsec:ds}

\noindent\textbf{Action selection.} 
The 44 manipulating actions of the \datasetname{} dataset were selected as follows: \textbf{(i)}~A list of candidate verb-object pairs was constructed by combining top verbs and objects sorted by the sum of their concreteness score \cite{brysbaert2014concreteness}. \textbf{(ii)}~Verb-object pairs corresponding to a visual change of an object state were manually selected. \textbf{(iii)}~Too general pairs were manually removed from the list, \eg. \textit{open a door}. \textbf{(iv)}~Similar or consecutive actions were joined into a single verb-object category, \eg \textit{cut} and \textit{peel an avocado}. \textbf{(v)}~YouTube API was queried for videos corresponding to a search term ``How to \textit{verb} an \textit{object}?'', ``\textit{verb} an \textit{object}'' or similar. \textbf{(vi)}~Categories with a small number of videos were removed.

\paragraphcustom{Action-state annotation.}
We hold out a small fraction of videos and manually annotate them for evaluation.
For each state-changing action, 30 videos are randomly sampled and annotated.
As the videos are uncurated, some of them do not contain the object nor the action of interest. 
Thus, not all the held-out videos are exhaustively annotated with states and actions.

Each video is divided into one second time intervals and each interval is assigned one of the following labels: \textit{background}, \textit{initial state}, \textit{action}, \textit{end state}.
We assign the \textit{initial} (resp. \textit{end}) \textit{states} to frames containing an object of interest that is visually similar to its appearance right at the beginning (resp. end) of the manipulating action.
The \textit{background} label is used when the object of interest is not clearly visible within the time interval.
In total, 667 videos with combined duration of 48 hours were annotated, yielding 15 videos per state-changing action on average.
Given the ratio between annotated and deliberately unannotated videos processed by our annotators, we estimate that approximately half of the videos in the dataset are \textit{noisy} in the sense that either the object, action or both are not clearly visible in the video.
The proportion of annotated labels in the test set is the following: \textit{initial state} 5\%, \textit{end state} 12\%, \textit{action} 42\% with the rest being labelled as \textit{background}.

\paragraphcustom{Video statistics.} Figure \ref{fig:ds_stats} shows the number of videos in our \datasetname{} data\-set for given video lengths. Only less than 15\% of the videos are shorter than one minute, and almost 60\% of the videos are longer than three minutes. On average, \textit{cherry pitting} has the shortest videos with the average duration of 2.6 minutes, the longest videos on average are in \textit{outlet installing} class with the mean duration of 7.2 minutes. The number of videos in each class varies from 265 in \textit{juice pouring} to 1914 in \textit{tortilla wrapping}. The list of all dataset classes is shown in Table \ref{tab:per-class}.

\begin{figure}[t]
    \centering
    \includegraphics[width=1.0\linewidth]{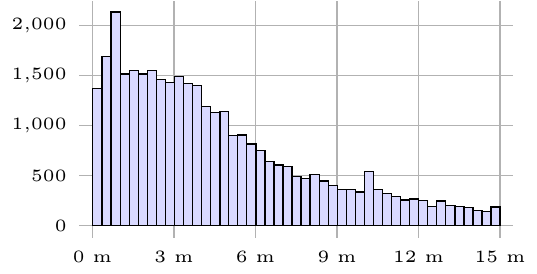}
    \caption{Histogram of video lengths in our \datasetname{} dataset.}
    \label{fig:ds_stats}
\end{figure}

\section{Different feature extractors}\label{supsec:featAblation}
In our model we use 2D ResNeXT pre-trained on ImageNet-21K \cite{OurResnextWeights} and 3D TSM ResNet50 pre-trained on HowTo100M and AudioSet \cite{Alayrac2020MMVN} as feature extractors. Besides those, we also tested CLIP \cite{clip} model and S3D trained using MIL-NCE loss on HowTo100M dataset \cite{miech2019end2end}. We show results of the tested models in Table \ref{tab:featAblationTable}. In case of CLIP, we observe substantial improvements in our metrics can be made when the last projection layer is removed ($\dagger$). We can also see that TSM-ResNet features contain additional information and improve action precision when used jointly with both ResNeXT and CLIP, even though their individual performance is low. On the contrary, using jointly only image extractors does not yield any benefit in action detection.

\begin{table}
  \centering
  {\small
  \begin{tabular}{cl@{~~}c@{~~}c@{~~}}
    \toprule
    Type & Feature extractor & St prec. & Ac prec. \\
    \midrule
    \multirow{2}{*}{Video} & MIL-NCE S3D \cite{miech2019end2end} & 0.22 & 0.49 \\
    & TSM-ResNet \cite{Alayrac2020MMVN} & 0.25 & 0.56 \\
    \midrule
    \multirow{4}{*}{Image} & CLIP ViT-B/16 \cite{clip} & 0.33 & 0.61 \\
    & CLIP ViT-B/16\textsuperscript{$\dagger$} \cite{clip} & \textbf{0.35} & 0.63 \\
    & ResNeXT \cite{OurResnextWeights} & 0.33 & 0.66 \\
    & ResNeXT + CLIP ViT-B/16\textsuperscript{$\dagger$} & 0.34 & 0.66 \\
    \midrule
    \multirow{3}{*}{Both} & CLIP ViT-B/16\textsuperscript{$\dagger$} + TSM-ResNet & 0.34 & 0.66 \\
    & ResNeXT + MIL-NCE S3D & 0.34 & 0.66 \\
    & ResNeXT + TSM-ResNet & \textbf{0.35} & \textbf{0.68} \\
    \bottomrule
    \multicolumn{3}{l}{\footnotesize \textsuperscript{$\dagger$} Without the last projection layer.}
  \end{tabular}
  }
  \caption{Comparison of different feature extractors in our method.}
  \label{tab:featAblationTable}
\end{table}

\section{Additional results}\label{supsec:results}

In this section we report per-class results, show additional examples and provide a qualitative analysis. % of the results.  

\paragraphcustom{Per-class results.} In Table \ref{tab:per-class} we report action and state precision for all dataset classes individually. We compare results of our approach to: (i) the state-of-the-art method for learning object states and state-modifyng actions by Alayrac \etal \cite{Alayrac16ObjectStates} as well as (ii) the zero-shot CLIP baseline~\cite{clip}. 
Details of the experimental set-up for both these baseline methods are in \ARXIVversion{Section \ref{subsec:comparison}}{Section 5.4 in the main paper}.

\paragraphcustom{Qualitative results.} Figures \ref{fig:res1}, \ref{fig:res2}, \ref{fig:res4} and \ref{fig:res3} show qualitative results for a set of selected classes. Figures \ref{fig:res1}, \ref{fig:res2} and \ref{fig:res4} illustrate predictions for classes where the learning of object state and action classifiers is successful. Figure \ref{fig:res3} shows the main limitations of our method, as described in Limitations in \ARXIVversion{Section \ref{subsec:comparison}}{Section 5.4 in the main paper}.
Example videos for each class were chosen as those with the highest prediction scores (within their class), where the prediction score is defined as:
\begin{align}
	\max_{l\in\mathcal{D}_v} \ & h_1(x_{l_{s_1}})\cdot g(x_{l_a}) \cdot h_2(x_{l_{s_2}}),
\end{align}
where $h_1(x_{l_{s_1}})$ is the classifier score of the initial state, $g(x_{l_a})$ is the classifier score of the action, and $h_2(x_{l_{s_2}})$ is the classifier score of the end state.  
We do not show videos with clearly visible faces, videos of poor quality or uninteresting static videos. Therefore, the shown videos may not be the highest ranked but (up to) fourth or fifth in the list of highest scoring videos for a given class.

\paragraphcustom{Analysis of the results.} As shown in Table~\ref{tab:per-class} as well as Figures~\ref{fig:res1}, \ref{fig:res2}, \ref{fig:res4} and \ref{fig:res3}, there can be large differences in performance between individual classes. We attribute these differences to the fact that some actions and object states are visually clearly defined whereas some can be visually ambiguous. Here are some of our qualitative findings:
\begin{description}[labelwidth=.5em, leftmargin=!, itemsep=0em]
    \item \textbf{(a) Peeling, slicing, chopping, cutting.} We observe strong performance for peeling or cutting any type of food. The actions and the states are visually distinct. The action has clear start and end, it is accompanied by visually distinct objects such as a knife. Also, there are many how-to videos for these actions.
    \item \textbf{(b) Frying, grinding, melting.} The initial and the end states of objects manipulated by actions such as frying or melting are visually distinct but the action itself is long without clearly defined start and end. For example the causal start of the melting action is turning on the heat source, which requires deep context understanding and even may not be shown in a video.
    \item \textbf{(c) Whisking, rolling, cleaning, tying.} For whisking or rolling, the action is clearly defined by an accompanying object such as a whisk or a roller but the visual difference between the object states can be rather small. For example, a cream is still white no matter the action, a dough is only a bit thinner, \etc.
    
    \item \textbf{(d) Opening, pouring.} There are not that many YouTube instructional videos for primitive actions such as pouring, thus the videos of these categories contain large variety of sub-actions, advertisement clips, and other noise making it much harder to correctly discover the states and the action.
    
    \item \textbf{(e) Drilling, wrapping, cutting.} These classes often have clearly defined actions and visually distinct states, but there is a large variance of appearance of the initial and the end state across different videos. For example, in some videos only a single hole is drilled but in other videos five new holes are drilled next to a handful of existing ones.
    
    \item \textbf{(f) Inflating, pitting, starting, removing.} For some of these classes, it is difficult to pinpoint the exact location of the action in the video. Also, the visual difference between the initial and the end state can be quite small. 
    For example, a partially deflated ball can be often recognized only by touch, not by vision.
\end{description}
We believe that additional forms of supervision, such as incorporating the audio signal or the language narration, may be needed to learn some of the hardest object changes.

\ifarxiv
  % force table to be at the top
  \makeatletter
  \setlength{\@fptop}{0pt}
  \makeatother
\fi

\begin{table}[t]
\centering

{\small\footnotesize
\begin{tabular}{lc@{~~}c@{~~}c|c@{~~}c@{~~}c@{~~}}
\toprule
\multirow{2}{*}{State-modifying action} & \multicolumn{3}{c|}{St prec.} & \multicolumn{3}{c}{Ac prec.} \\
& Ours & \cite{Alayrac16ObjectStates} & \cite{clip} & Ours & \cite{Alayrac16ObjectStates} & \cite{clip}\\
\midrule
\multicolumn{7}{l}{\textbf{(a)} \textit{Visually distinct states and actions}} \\
Apple Peeling/Cutting        & \textbf{0.46} & {0.41} & {0.41} & {0.79} & {0.69} & \textbf{0.81} \\
Avocado Peeling/Slicing      & \textbf{0.44} & {0.38} & {0.28} & \textbf{0.92} & {0.88} & {0.88} \\
Beer Pouring                 & \textbf{0.37} & {0.22} & {0.25} & \textbf{0.79} & {0.62} & {0.44} \\
Corn Peeling                 & \textbf{0.51} & {0.45} & {0.19} & \textbf{0.68} & {0.57} & {0.48} \\
Dragon Fruit Peeling/Cutting & {0.47} & \textbf{0.48} & \textbf{0.48} & \textbf{0.94} & {0.90} & \textbf{0.94} \\
Eggs Peeling                 & {0.36} & {0.25} & \textbf{0.38} & \textbf{0.60} & {0.25} & {0.50} \\
Garlic Peeling/Chopping      & \textbf{0.44} & {0.39} & {0.28} & \textbf{1.00} & {0.89} & \textbf{1.00} \\
Onions Peeling/Chopping      & \textbf{0.49} & {0.40} & {0.23} & \textbf{0.91} & {0.73} & {0.80} \\
Paper Plane Folding          & {0.37} & \textbf{0.38} & {0.21} & \textbf{1.00} & \textbf{1.00} & \textbf{1.00} \\
Pineapple Peeling/Slicing    & {0.28} & \textbf{0.33} & {0.12} & \textbf{1.00} & {0.90} & \textbf{1.00} \\
T-shirt Dyeing               & \textbf{0.48} & {0.47} & {0.35} & {0.86} & \textbf{1.00} & {0.82} \\
Tortilla Wrapping            & {0.39} & {0.27} & \textbf{0.54} & \textbf{0.71} & {0.53} & {0.53} \\
\midrule
\multicolumn{7}{l}{\textbf{(b)} \textit{Visually distinct states, actions with unclear boundaries}} \\
Bacon Frying                 & \textbf{0.40} & {0.22} & {0.25} & {0.20} & {0.39} & \textbf{0.61} \\
Chocolate Melting            & \textbf{0.54} & {0.32} & {0.47} & \textbf{0.65} & {0.35} & {0.59} \\
Coffee Grinding              & {0.47} & {0.29} & \textbf{0.50} & {0.25} & \textbf{0.50} & {0.17} \\
Potatoes Frying              & \textbf{0.32} & {0.26} & {0.27} & \textbf{0.98} & {0.96} & {0.95} \\
\midrule
\multicolumn{7}{l}{\textbf{(c)} \textit{Visually distinct actions, small visual changes between states}} \\
Cake Frosting                & {0.22} & \textbf{0.25} & {0.19} & {0.50} & {0.42} & \textbf{0.79} \\
Cream Whipping               & {0.36} & \textbf{0.39} & {0.30} & {0.42} & \textbf{0.50} & {0.32} \\
Dough Rolling                & {0.27} & {0.27} & \textbf{0.37} & {0.62} & {0.60} & \textbf{0.73} \\
Eggs Whisking                & \textbf{0.30} & {0.29} & {0.25} & \textbf{0.86} & {0.64} & {0.64} \\
Fish Filleting               & {0.21} & {0.22} & \textbf{0.23} & {0.87} & {0.90} & \textbf{0.95} \\
Pan Cleaning                 & {0.46} & \textbf{0.53} & {0.36} & \textbf{0.96} & {0.78} & {0.94} \\
Rubik's Cube Solving         & \textbf{0.16} & {0.14} & {0.03} & {0.91} & {0.67} & \textbf{1.00} \\
Shoes Cleaning               & {0.18} & {0.21} & \textbf{0.26} & \textbf{0.90} & {0.84} & {0.89} \\
Tie Tying                    & {0.46} & \textbf{0.50} & {0.12} & \textbf{1.00} & \textbf{1.00} & \textbf{1.00} \\
Ribbon/Bow Tying             & \textbf{0.28} & {0.17} & {0.19} & \textbf{0.98} & {0.94} & {0.89} \\
Rope/Knot Tying              & \textbf{0.36} & {0.25} & {0.29} & \textbf{0.83} & \textbf{0.83} & {0.58} \\
\midrule
\multicolumn{7}{l}{\textbf{(d)} \textit{Not many how-to videos available}} \\
Butter Melting               & {0.22} & \textbf{0.25} & {0.17} & \textbf{0.67} & {0.50} & {0.50} \\
Candle Lighting              & {0.50} & {0.31} & \textbf{0.62} & \textbf{0.29} & {0.12} & {0.00} \\
Champagne Opening            & \textbf{0.45} & {0.39} & {0.36} & \textbf{0.24} & {0.14} & {0.14} \\
Juice Pouring                & {0.32} & {0.32} & \textbf{0.45} & {0.12} & {0.00} & \textbf{0.36} \\
Milk Boiling                 & {0.28} & {0.26} & \textbf{0.35} & {0.21} & \textbf{0.29} & {0.24} \\
Milk Pouring                 & {0.20} & \textbf{0.40} & {0.20} & {0.50} & {0.30} & \textbf{0.70} \\
Tea Pouring                  & \textbf{0.17} & {0.08} & \textbf{0.17} & \textbf{0.39} & {0.00} & {0.17} \\
\midrule
\multicolumn{7}{l}{\textbf{(e)} \textit{Distinct states and actions but high variance in appearance}} \\
Gift/Box Wrapping            & {0.20} & {0.21} & \textbf{0.26} & {0.84} & \textbf{0.95} & \textbf{0.95} \\
Outlet Installing            & {0.23} & {0.15} & \textbf{0.35} & \textbf{0.87} & {0.77} & {0.69} \\
Pancake Flipping             & \textbf{0.36} & {0.31} & {0.29} & \textbf{0.19} & \textbf{0.19} & {0.14} \\
Tile Cutting                 & {0.28} & {0.35} & \textbf{0.45} & \textbf{0.63} & {0.50} & {0.60} \\
Tree Cutting                 & \textbf{0.40} & {0.19} & {0.22} & \textbf{0.70} & {0.61} & {0.56} \\
Wood Drilling                & {0.29} & {0.14} & \textbf{0.41} & {0.45} & {0.36} & \textbf{0.73} \\
\midrule
\multicolumn{7}{l}{\textbf{(f)} \textit{Minimal visual change of states, actions with unclear boundaries}} \\
Ball Inflating               & \textbf{0.22} & {0.20} & {0.05} & {0.40} & \textbf{0.60} & {0.30} \\
Cherries Pitting             & {0.31} & {0.25} & \textbf{0.38} & \textbf{0.50} & {0.12} & \textbf{0.50} \\
Grill Starting               & \textbf{0.33} & {0.12} & \textbf{0.33} & \textbf{0.83} & {0.58} & {0.58} \\
Weed Removing                & {0.33} & {0.41} & \textbf{0.45} & \textbf{0.70} & {0.64} & {0.55} \\
\midrule
Average                     & \textbf{0.35} & {0.30} & {0.30} & \textbf{0.68} & {0.59} & {0.63} \\
\bottomrule
\end{tabular}
}

\caption{{\bf Per-class state and action precision} on our  \datasetname{} dataset. Our approach improves on average over the state-of-the-art approach of~\cite{Alayrac16ObjectStates} and CLIP ViT-L/14 \cite{clip}. }  
\label{tab:per-class}
\end{table}

\section{Potential negative societal impact}\label{supsec:impact}

Our work is based on models trained without human annotation. However, the models are still subject to biases in the training data. We gather our training dataset from the YouTube platform, which makes our results dependent on the availability and the quality of the videos uploaded to the site. In addition, the content on the platform is potentially not uniformly distributed across countries, ages, ethnic groups, \etc. Thus our models can under-perform for some actions, for example, conducted only by minorities. Also, if an action has multiple possible realizations, there is a risk of learning only the realization that is the most prevalent in the data.

\ifarxiv
  % reset the figure position to the default value
  \clearpage
  \makeatletter
  \setlength{\@fptop}{0pt plus 1.0fil}
  \makeatother
\fi

\begin{figure*}[t]
    \centering
    \begin{tabular}{c|c}
    \textbf{Paper plane folding} & \textbf{Ribbon tying} \\[3pt]
    \includegraphics[width=0.47\linewidth]{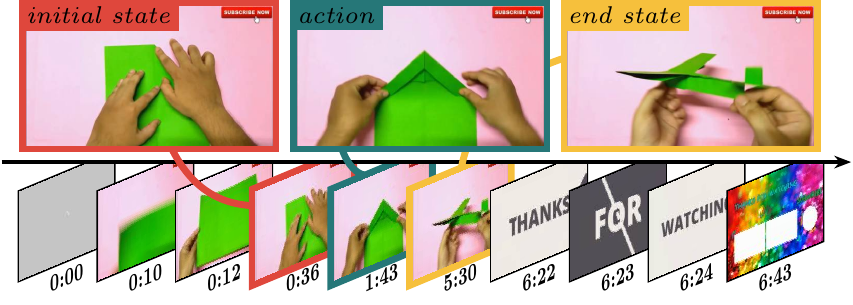} & \includegraphics[width=0.47\linewidth]{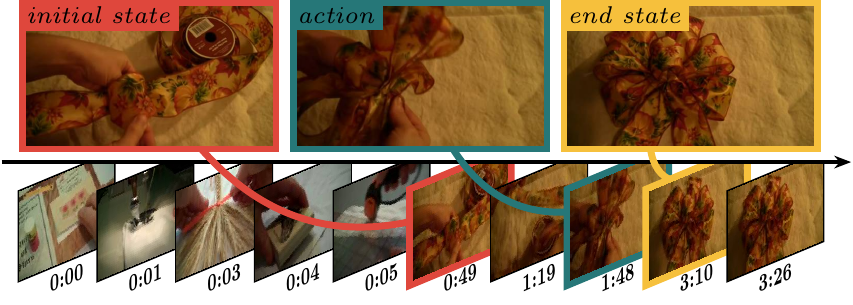} \\
    \includegraphics[width=0.47\linewidth]{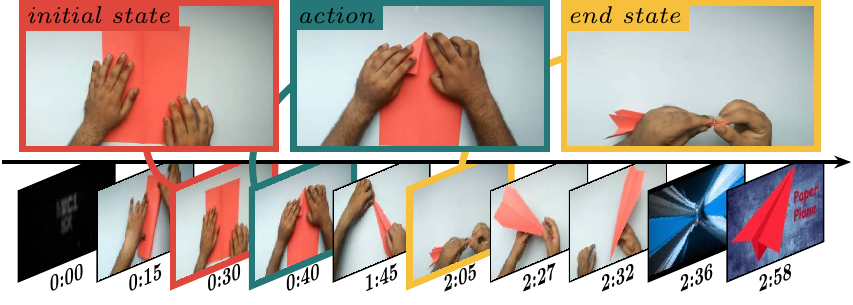} & \includegraphics[width=0.47\linewidth]{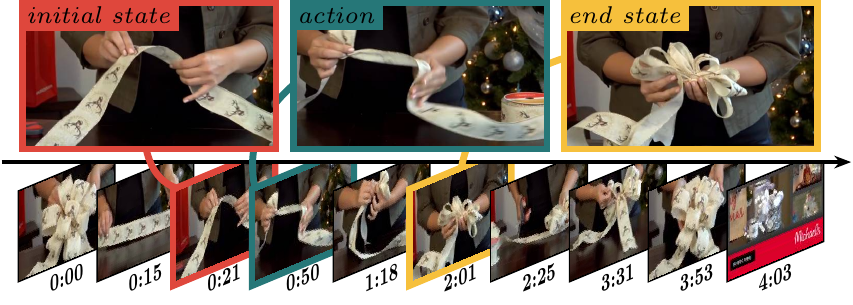} \\
    \includegraphics[width=0.47\linewidth]{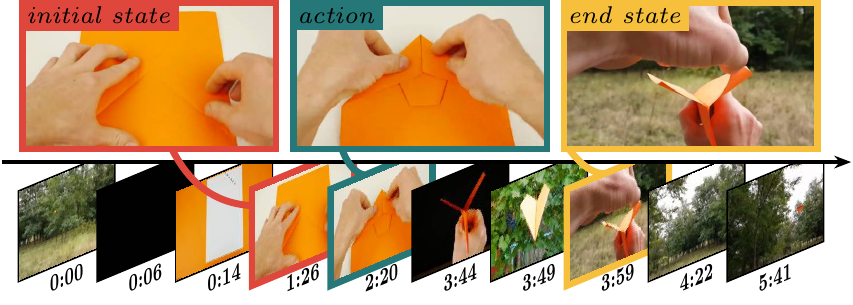} & \includegraphics[width=0.47\linewidth]{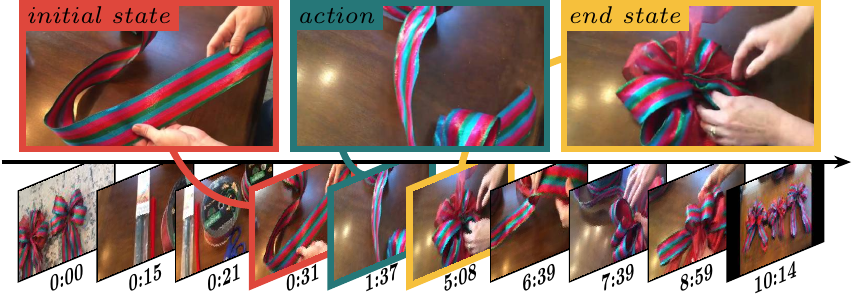} \\
    \hline \\[-8pt]
    \textbf{T-shirt dyeing} & \textbf{Outlet installing} \\[3pt]
    \includegraphics[width=0.47\linewidth]{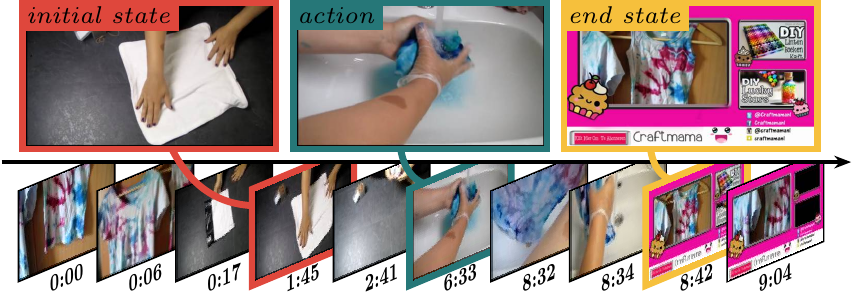} & \includegraphics[width=0.47\linewidth]{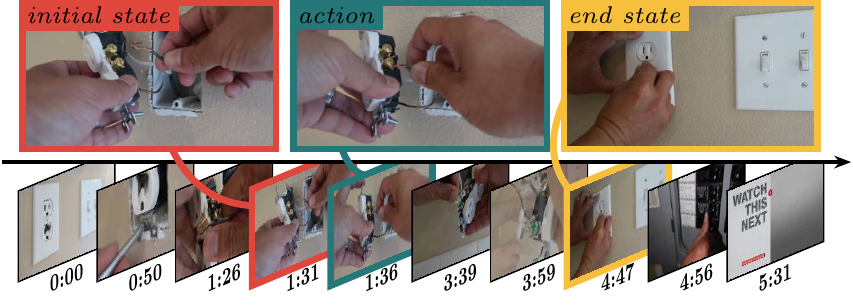} \\
    \includegraphics[width=0.47\linewidth]{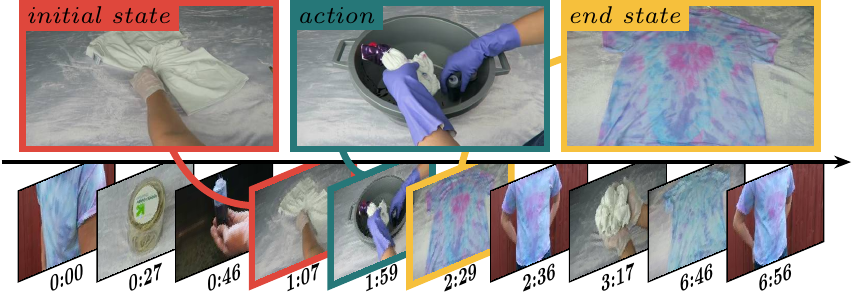} & \includegraphics[width=0.47\linewidth]{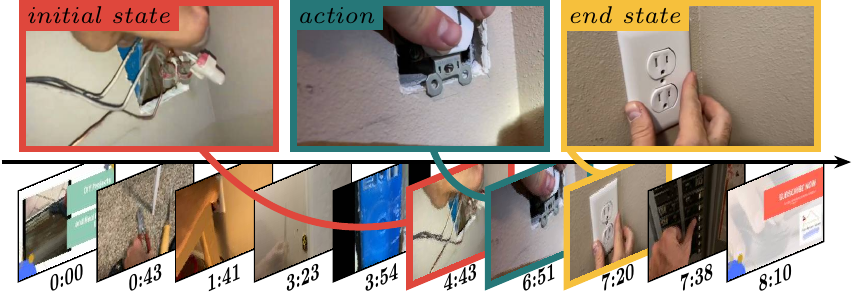} \\
    \includegraphics[width=0.47\linewidth]{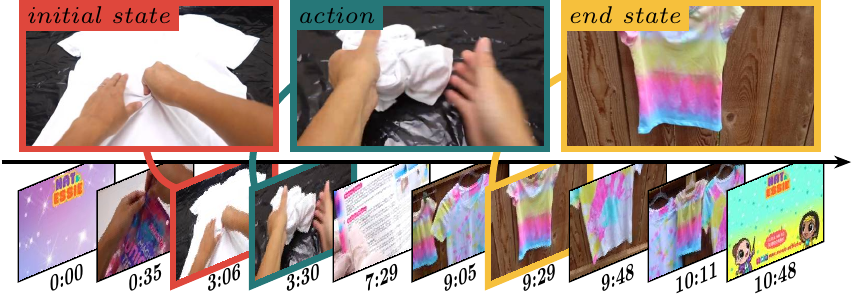} & \includegraphics[width=0.47\linewidth]{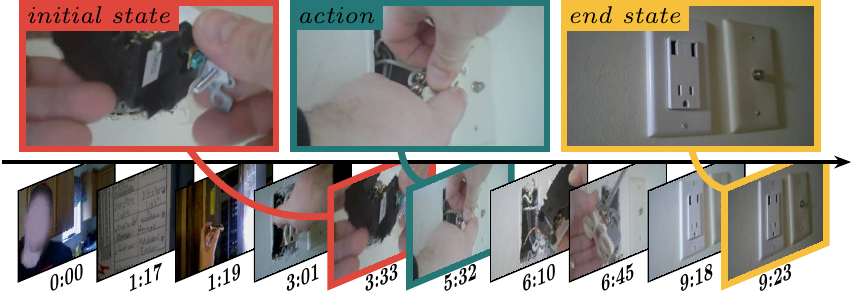} \\
    \end{tabular}
    \caption{\textbf{Additional example results for four different DIY classes, ``Paper plane folding", ``Ribbon tying", ``T-shirt dyeing" and ``Outlet installing".} For each class we show the temporal localization of the initial state, state-modifying action, and the end state in three different example videos. Note how our model is able to learn the object states and the state-modifying action despite the large appearance variation in the videos (viewpoint, environment, intra-class variation for both the object states and the action). }
    \label{fig:res1}
\end{figure*}

\begin{figure*}[t]
    \centering
    \begin{tabular}{c|c}
    \textbf{Chocolate melting} & \textbf{Fish filleting} \\[3pt]
    \includegraphics[width=0.47\linewidth]{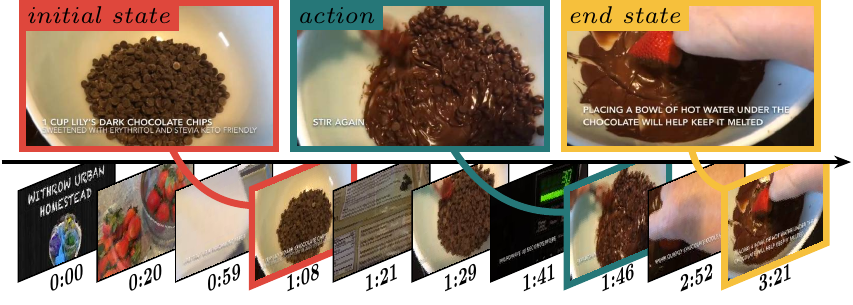} & \includegraphics[width=0.47\linewidth]{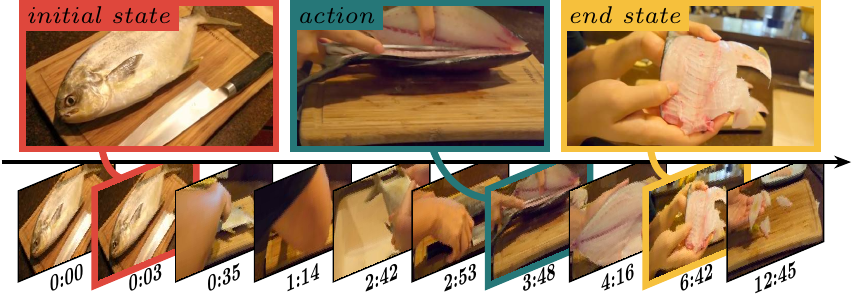} \\
    \includegraphics[width=0.47\linewidth]{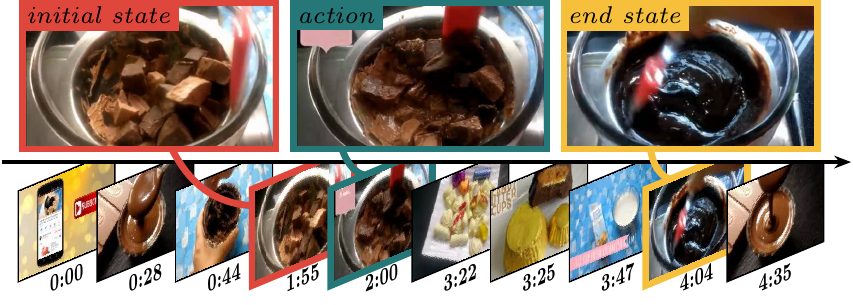} & \includegraphics[width=0.47\linewidth]{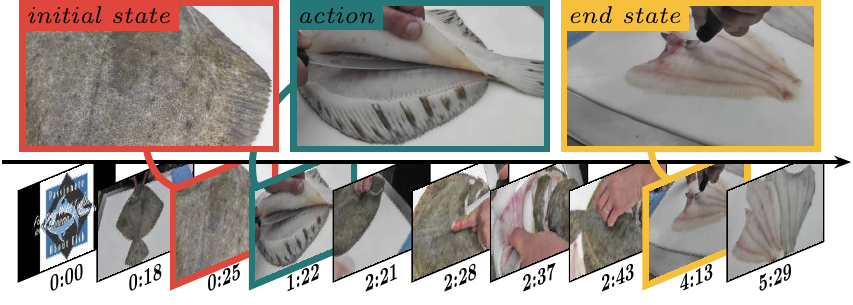} \\
    \includegraphics[width=0.47\linewidth]{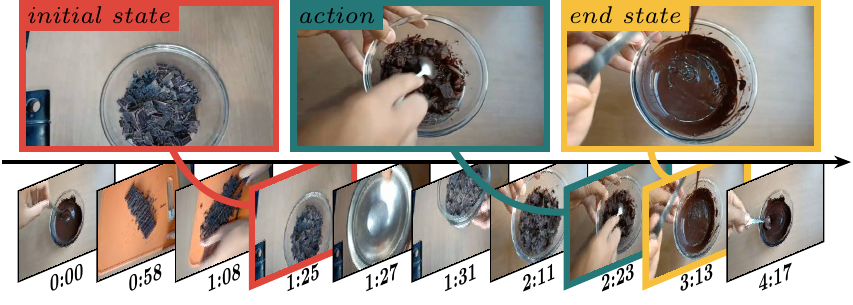} & \includegraphics[width=0.47\linewidth]{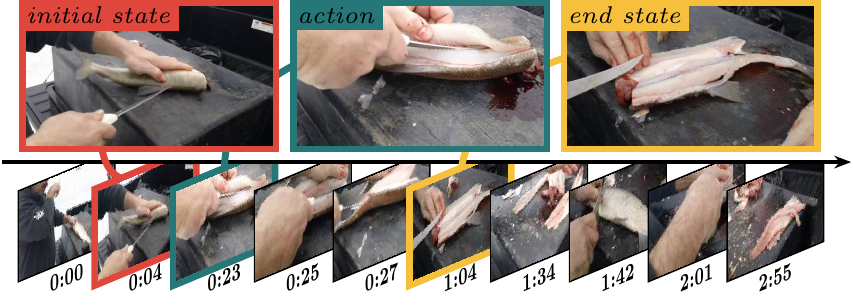} \\
    \hline \\[-8pt]
    \textbf{Dragon fruit peeling} & \textbf{Bacon frying} \\[3pt]
    \includegraphics[width=0.47\linewidth]{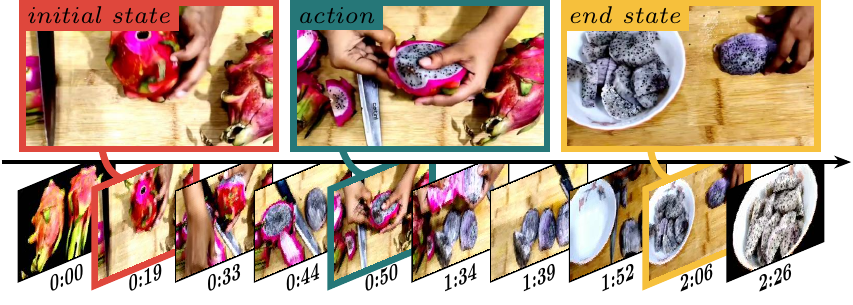} & \includegraphics[width=0.47\linewidth]{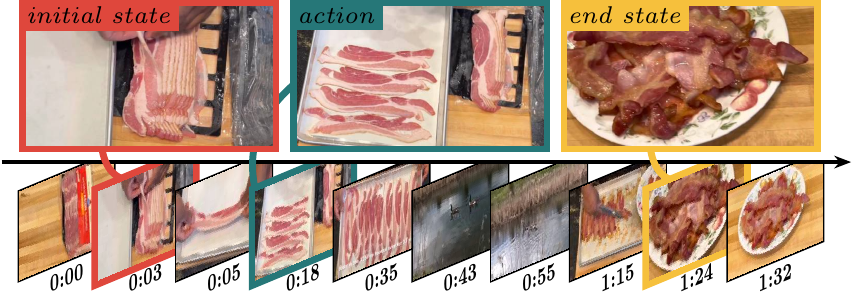} \\
    \includegraphics[width=0.47\linewidth]{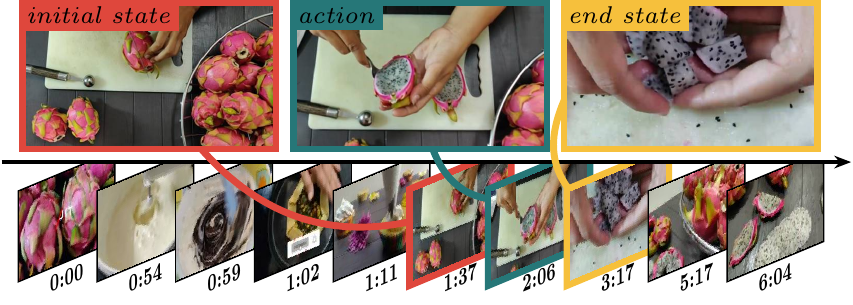} & \includegraphics[width=0.47\linewidth]{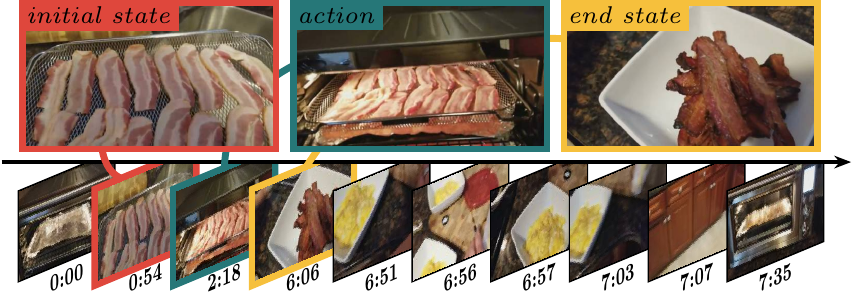} \\
    \includegraphics[width=0.47\linewidth]{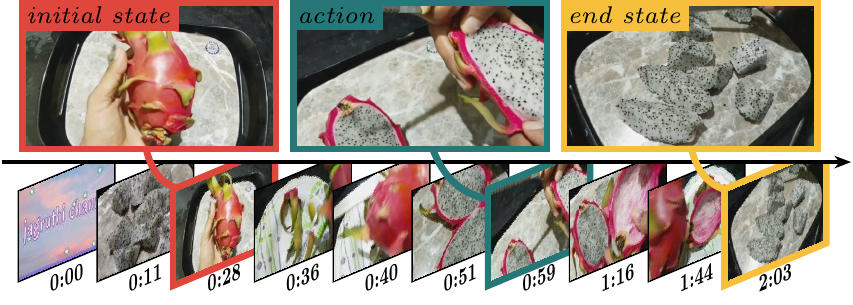} & \includegraphics[width=0.47\linewidth]{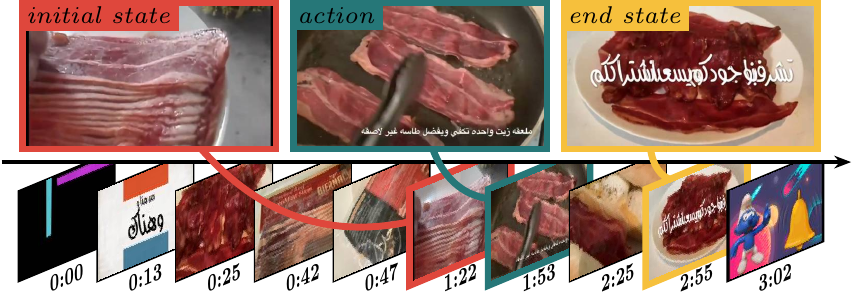} \\
    \end{tabular}
    \caption{\textbf{Additional example results for four different meal preparation classes, ``Chocolate melting", ``Fish filleting", ``Dragon fruit peeling" and ``Bacon frying".} For each class we show the temporal localization of the initial state, state-modifying action, and the end state in three different example videos. Note how our model is able to learn the object states and the state-modifying action despite the large appearance variation in the videos  (viewpoint, environment, intra-class variation for both the object states and the action).  }
    \label{fig:res2}
\end{figure*}

\begin{figure*}[t]
    \centering
    \begin{tabular}{c|c}
    \textbf{Cream whipping} & \textbf{Beer pouring} \\[3pt]
    \includegraphics[width=0.47\linewidth]{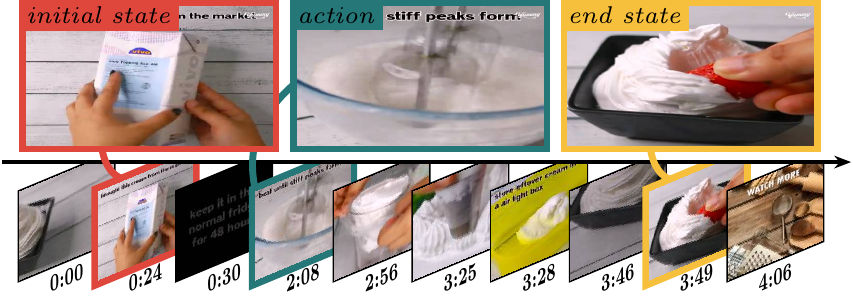} & \includegraphics[width=0.47\linewidth]{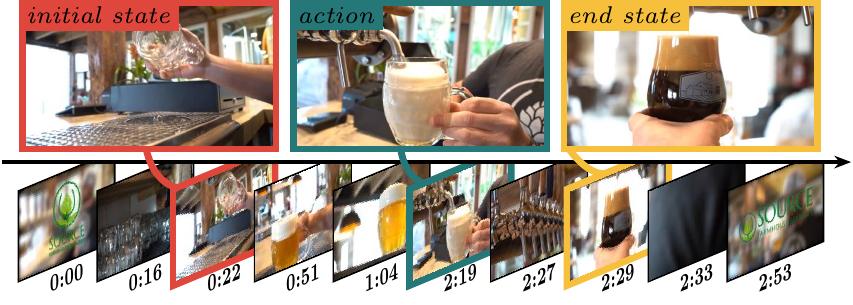} \\
    \includegraphics[width=0.47\linewidth]{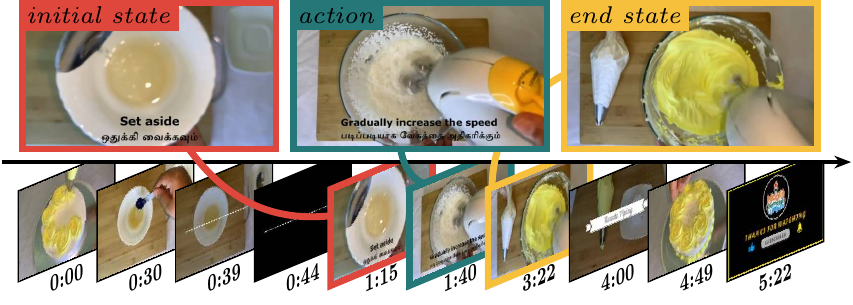} & \includegraphics[width=0.47\linewidth]{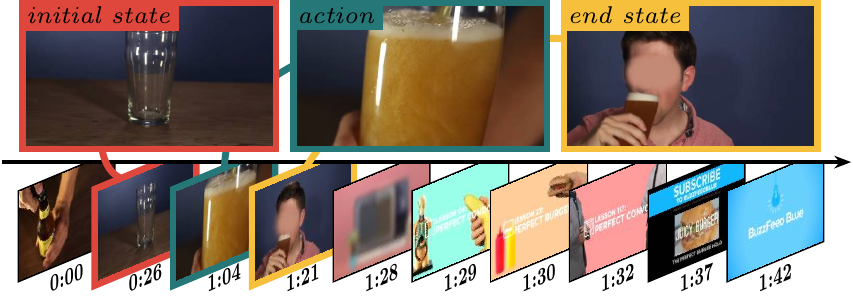} \\
    \includegraphics[width=0.47\linewidth]{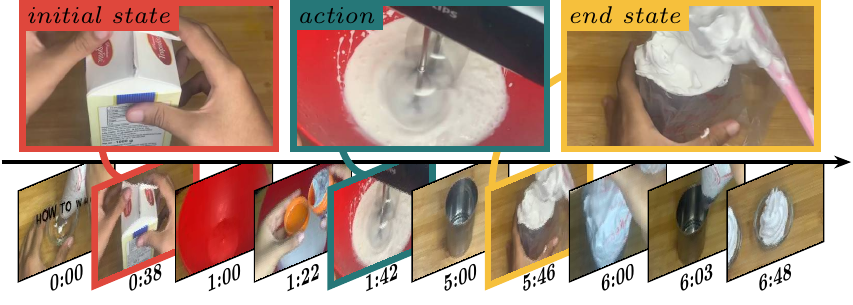} & \includegraphics[width=0.47\linewidth]{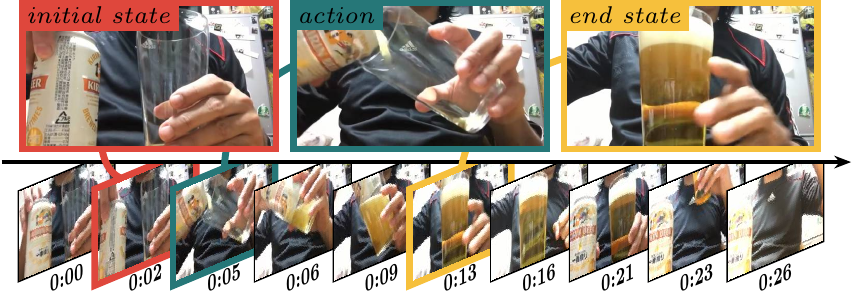} \\
    \hline \\[-8pt]
    \textbf{Pan cleaning} & \textbf{Rubik's cube solving} \\[3pt]
    \includegraphics[width=0.47\linewidth]{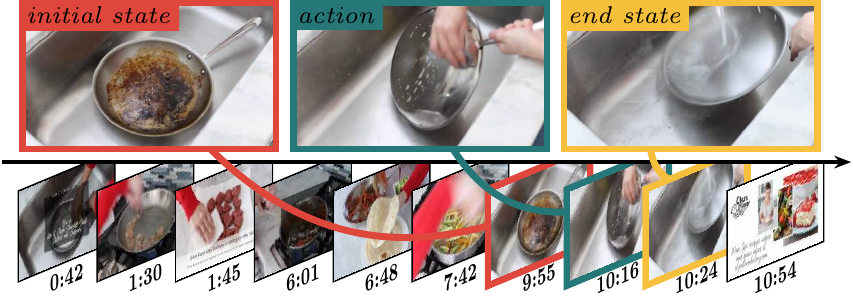} & \includegraphics[width=0.47\linewidth]{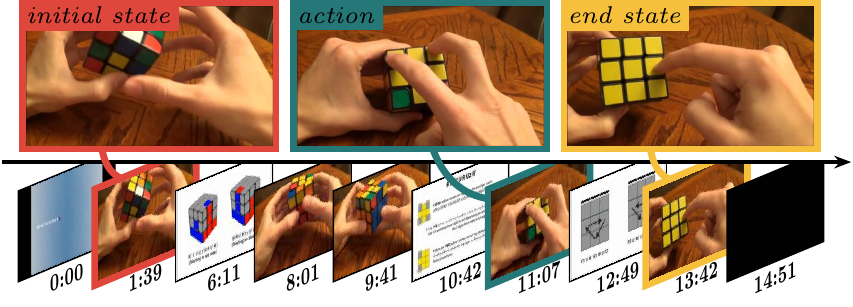} \\
    \includegraphics[width=0.47\linewidth]{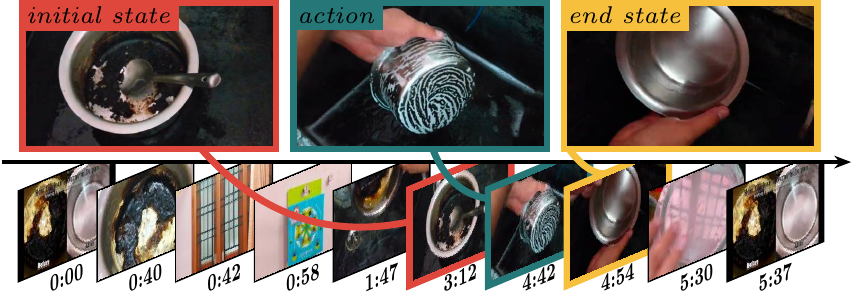} & \includegraphics[width=0.47\linewidth]{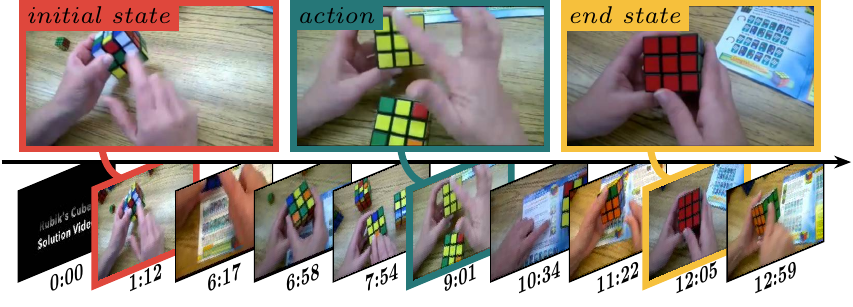} \\
    \includegraphics[width=0.47\linewidth]{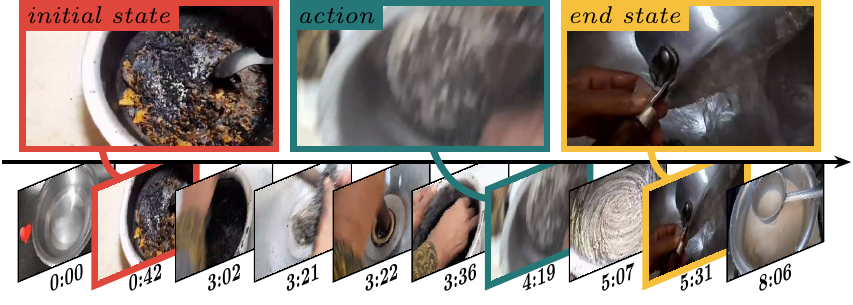} & \includegraphics[width=0.47\linewidth]{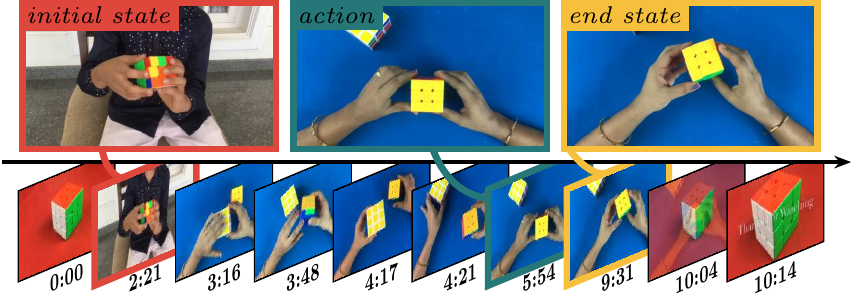} \\
    \end{tabular}
    \caption{\textbf{Additional example results for four different classes, ``Cream whipping", ``Beer pouring", ``Pan cleaning" and ``Rubik's cube solving".} For each class we show the temporal localization of the initial state, state-modifying action, and the end state in three different example videos. Note how our model is able to learn the object states and the state-modifying action despite the large appearance variation in the videos  (viewpoint, environment, intra-class variation for both the object states and the action).  }
    \label{fig:res4}
\end{figure*}

\begin{figure*}[t]
    \centering
    \begin{tabular}{c|c}
    \textbf{(a) Onion chopping} & \textbf{(b) Tea pouring} \\[3pt]
    \includegraphics[width=0.47\linewidth]{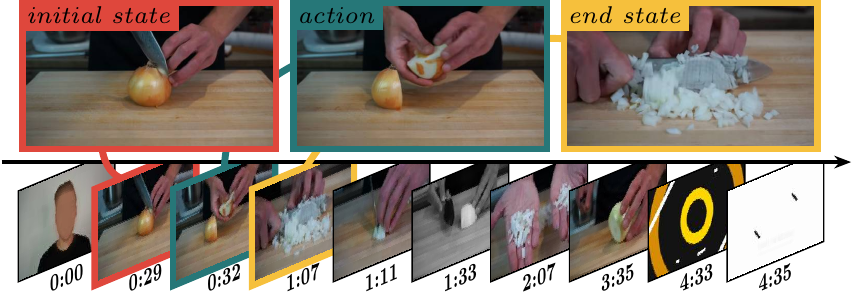} & \includegraphics[width=0.47\linewidth]{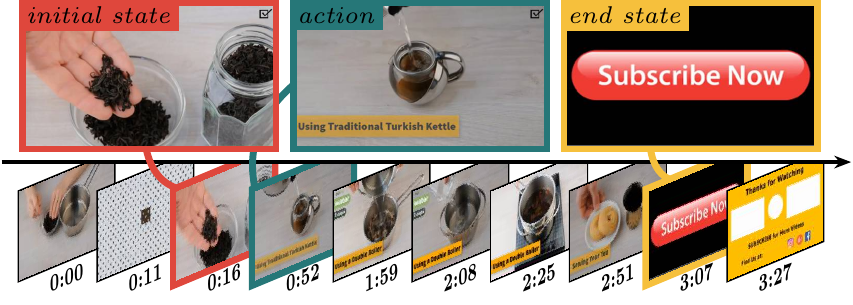} \\
    \includegraphics[width=0.47\linewidth]{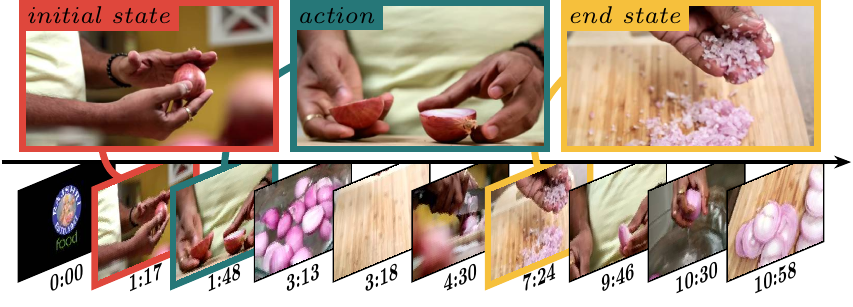} & \includegraphics[width=0.47\linewidth]{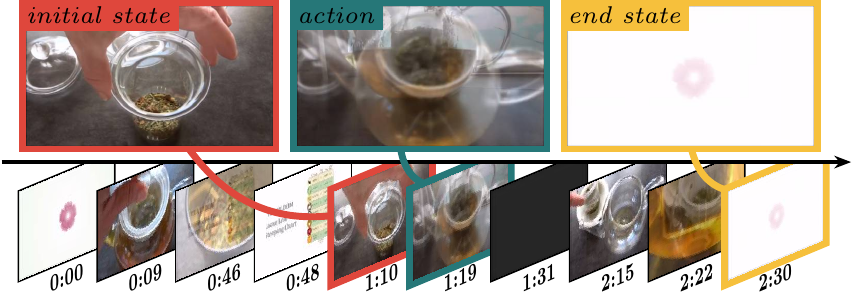} \\
    \includegraphics[width=0.47\linewidth]{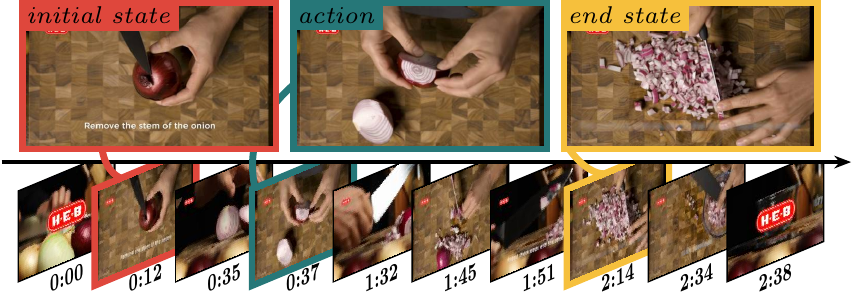} & \includegraphics[width=0.47\linewidth]{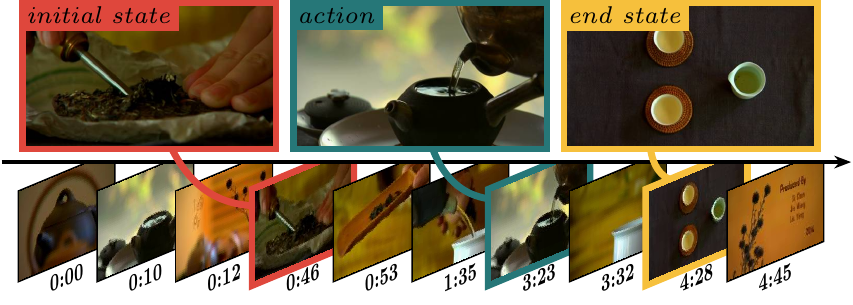} \\
    \hline \\[-8pt]
    \textbf{(c) Weed removing} & \textbf{(d) Butter melting} \\[3pt]
    \includegraphics[width=0.47\linewidth]{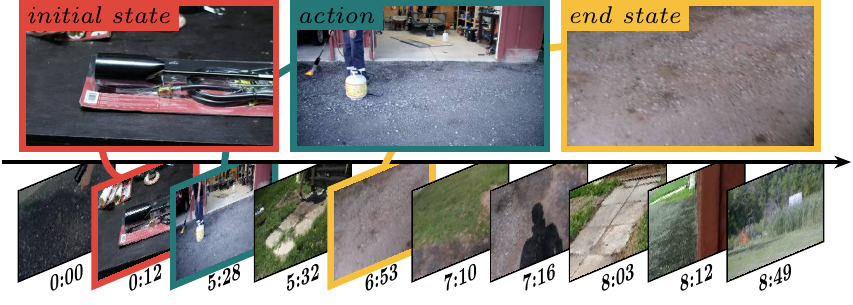} & \includegraphics[width=0.47\linewidth]{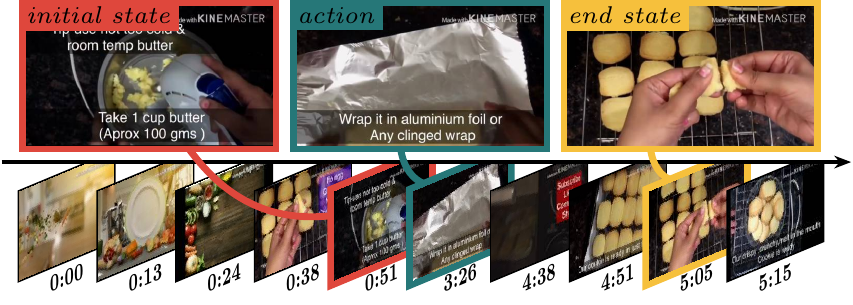} \\
    \includegraphics[width=0.47\linewidth]{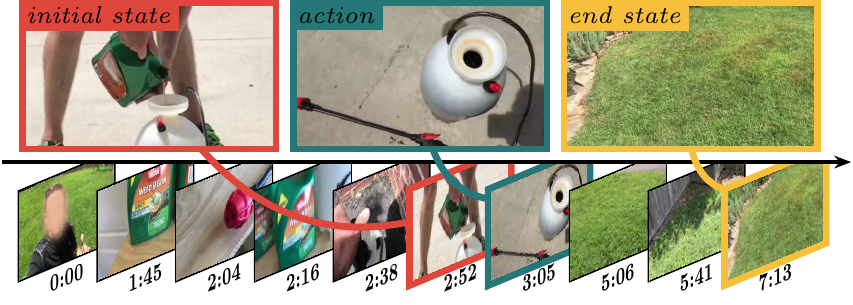} & \includegraphics[width=0.47\linewidth]{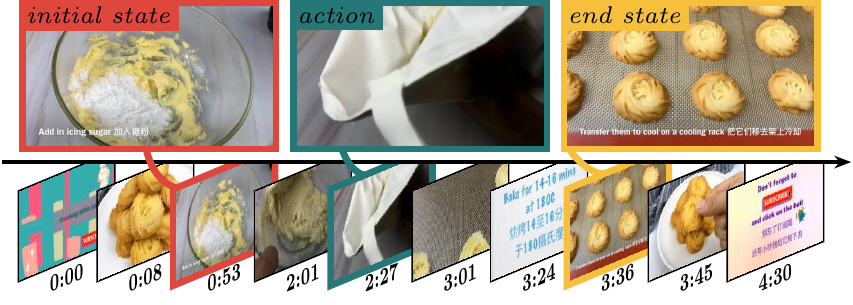} \\
    \includegraphics[width=0.47\linewidth]{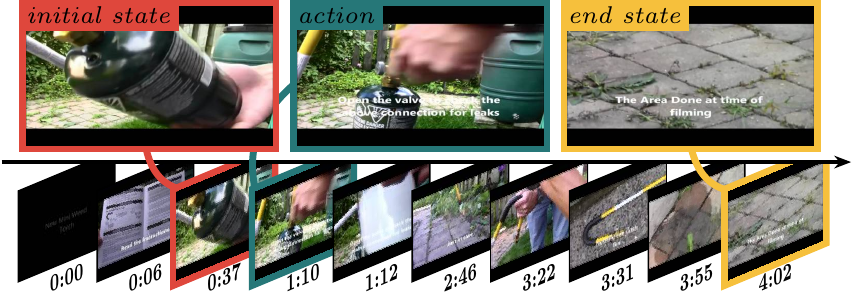} & \includegraphics[width=0.47\linewidth]{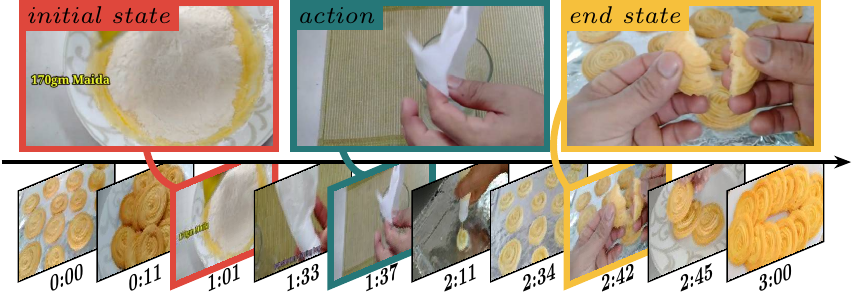} \\
    \end{tabular}
    \caption{\textbf{Examples of typical failure modes.} (a) The model learns a different action than the expected action (here \textit{holding a piece of onion} instead of \textit{chopping onion}). (b) The model discovers consistent visual appearance in the videos, which is just an artefact of the editing process (here single-colored frame at the and of a video as an end state). (c) Some categories can have a large variance in the appearance of the initial and the end state across different videos (here removing a single plant or clearing the whole path). (d) Some categories are not well represented on YouTube (here videos of \textit{making butter cookies} dominate the search results for query \textit{melting butter}).
    }
    \label{fig:res3}
\end{figure*}

}{}

\end{document}